%% file: main.tex
\documentclass[sigconf,nonacm]{acmart}

\AtBeginDocument{%
  }

\usepackage{algorithm}
\usepackage{algpseudocode}
\usepackage{multirow}
\usepackage{booktabs}
\usepackage{caption}
\usepackage{xspace}
\usepackage{makecell}
\usepackage{threeparttable}
\usepackage{float}
\usepackage{enumitem}

\begin{document}

\title{FAST: A Holistic Framework for Optimizing Memory-I/O, Computation, and Sampling in Temporal GNN Training}

\author{Yushu Cai}
\orcid{0009-0009-1878-744X}
\affiliation{%
  \institution{Xidian University}
  \city{Guangzhou}
  \state{Guangdong}
  \country{China}}
\email{yushucai@stu.xidian.edu.cn}

\author{Qingrui Zhu}
\affiliation{%
  \institution{Xidian University}
  \city{Guangzhou}
  \state{Guangdong}
  \country{China}}
\email{zhuqingrui@stu.xidian.edu.cn}

\author{Lei Liu}
\affiliation{%
  \institution{Xidian University}
  \city{Guangzhou}
  \state{Guangdong}
  \country{China}}
\email{liulei2303@stu.xidian.edu.cn}

\author{Kai Sheng}
\affiliation{%
  \institution{Xidian University}
  \city{Guangzhou}
  \state{Guangdong}
  \country{China}}
\email{kaisheng@xidian.edu.cn}

\author{Hao Chen}
\affiliation{%
  \institution{Hunan University}
  \city{Changsha}
  \state{Hunan}
  \country{China}}
\email{haochen@hnu.edu.cn}

\author{Xin He}
\affiliation{%
  \institution{Xidian University}
  \city{Guangzhou}
  \state{Guangdong}
  \country{China}}
\email{hexin@xidian.edu.cn}

\newcommand{\xin}[1]{\textcolor{black}{#1}}
\newcommand{\yushu}[1]{\textcolor{black}{#1}}

\settopmatter{printacmref=false} 
\renewcommand\footnotetextcopyrightpermission[1]{}
\pagestyle{plain}
\setlength{\abovecaptionskip}{2pt}
\setlength{\belowcaptionskip}{-2pt}
\setlength{\textfloatsep}{6pt plus 1pt minus 2pt}
\setlength{\floatsep}{4pt plus 1pt minus 2pt}

\setlength{\aboverulesep}{0pt}
\setlength{\belowrulesep}{0pt}
\captionsetup[table]{skip=2pt}

\begin{abstract}
\xin{Temporal Graph Neural Networks (TGNNs) are widely used for learning from dynamic graphs in applications such as recommendation, social network analysis, and traffic forecasting. However, scaling TGNN training to large dynamic graphs remains challenging due to three intertwined bottlenecks: memory I/O, irregular computation, and temporal neighbor sampling. Existing systems often optimize these stages in isolation, leaving substantial performance headroom on the table. We present \textsc{FAST}, a holistic framework that accelerates end-to-end TGNN training by jointly optimizing sampling, memory I/O, and computation. \textsc{FAST} introduces \emph{SlimCache}, which exploits within-batch compression and cross-batch caching to reduce host-device data movement under limited GPU memory budgets. It further designs thread-efficient graph operators tailored to sparse temporal subgraphs, improving GPU cache locality and reducing the latency of aggregation and edge softmax. In addition, \textsc{FAST} employs a topology-aware sampling strategy that improves CPU cache locality and accelerates temporal neighbor sampling. Extensive experiments on real-world large dynamic graphs show that \textsc{FAST} achieves an average of 2.1$\times$ (up to 4.7$\times$) speedup over state-of-the-art systems without sacrificing model accuracy.The code of FAST is publicly available at https://github.com/NoneBone/FAST.}
\end{abstract}


\thanks{Accepted to the 55th International Conference on Parallel Processing (ICPP 2026).}

\maketitle

\input{text/introduction}
\input{text/background}

\input{text/motivation}

\input{text/design}

\input{text/impl}

\input{text/evaluation}

\input{text/relatedWork}
\input{text/conclusion}

\bibliographystyle{ACM-Reference-Format}
\bibliography{references}

\appendix

\end{document}

%% file: text/introduction.tex
\section{Introduction}

\xin{Dynamic graphs naturally model timestamped interactions between real-world entities, with nodes representing entities and timestamped edges capturing their temporal relationships. Learning from such data is fundamental to applications including recommendation, social network analysis, traffic forecasting, and fraud detection \citep{Zhang2021DynamicGN, Jin2023SpatioTemporalGN}. Temporal Graph Neural Networks (TGNNs) \citep{tgn,tgat,sankar2020dysat} have become the dominant approach by capturing evolving message dynamics through recursive temporal message passing, combining temporal neighbor sampling with temporally encoded neighborhood aggregation to learn expressive, time-aware node representations. TGNNs excel at modeling evolving relationships, such as user interest drift in streaming services \citep{wiki2_paranjape2017motifs}, enabling accurate future behavior prediction.}

\xin{To scale TGNNs to large dynamic graphs, existing systems typically adopt a timestamp-ordered batch training paradigm \citep{tgn,tgat}. The input graph is partitioned into chronologically ordered batches, each containing a sequence of interactions. For each batch, the system (1) performs temporal neighbor sampling to construct a computation subgraph, (2) reads node states and feature data from host memory and transfers them to the GPU, and (3) executes forward and backward propagation to update representations and optimize model parameters. While effective in small-scale settings, this pipeline suffers from the following three severe  bottlenecks when scaled to large graphs.}

\xin{\textbf{Memory I/O Bottleneck.} A fundamental bottleneck is the massive data access overhead between host and GPU memory. Each mini-batch requires copying dynamically sampled subgraphs and associated features, saturating the host–device interconnect (e.g., 32 GB/s of PCIe 4.0 with 16 channels) and leaving GPU compute cores underutilized. In large-scale graphs, the memory I/O stage can dominate up to 78\% of total training time \citep{tgl,etc}. Existing frameworks attempt to mitigate this through feature compression (e.g., ETC \citep{etc}) or caching placement (e.g., SIMPLE \citep{simple}, TASER \citep{taser}), but their designs often ignore the heterogeneous redundancy patterns of nodes and edges, and are constrained by small GPU cache budgets.}

\xin{\textbf{Computation Bottleneck.} Beyond data movement, TGNN training suffers from inefficient computation on sparse dynamic graphs. The sampled subgraphs exhibit highly skewed degree distributions and irregular memory access patterns, leading to load imbalance and low cache utilization on GPUs \citep{Zhang2021DynamicGN, Jin2023SpatioTemporalGN}. Current work focuses on embedding reuse or redundancy elimination \citep{Li2023OrcaST, Wang2023TGOptRO} but overlooks the acceleration of core graph operators(aggregation and edge softmax), leaving the computation stage as a key bottleneck.}

\xin{\textbf{Sampling Bottleneck.} In addition, temporal neighbor sampling becomes a significant cost. CPU-based parallel samplers such as TGL \citep{tgl} are widely adopted, but they underutilize CPU cache hierarchies. GPU-based samplers (e.g., GNNFlow \citep{gnnflow}, MSpipe \citep{msPipe}, TASER \citep{taser}) achieve higher throughput \yushu{but with} custom implementations and poor reusability, making cross-framework comparisons difficult. This motivates a simple, high‑performance sampling strategy that improves CPU cache locality without sacrificing generality.}

\xin{In this paper, we present \textsc{FAST}, a holistic framework for optimizing memory I/O, computation, and sampling in TGNN training. \textsc{FAST} is built on the observation that TGNN training exhibits substantial within‑batch and cross‑batch redundancy in sampled subgraphs. We first design \emph{SlimCache}, which jointly exploits compression and caching to minimize host–device data movement under a limited GPU memory budget, carefully distinguishing node and edge reuse patterns. Second, we introduce thread‑efficient graph operators tailored to the sparsity and irregularity of temporal graphs, redesigning task partitioning and reduction paths to reduce latency in aggregation and edgeSoftmax. Third, we develop a topology‑aware sampling strategy that maps sampling threads to CPU cores based on subgraph similarity, improving cache locality and accelerating the sampling stage.}

In summary, our main contributions are as follows:
\begin{itemize}[leftmargin=10pt, itemindent=0pt, labelwidth=0pt, labelsep=0.5em, topsep=0pt, partopsep=0pt, itemsep=0pt]

\item We identify and characterize the joint bottlenecks of memory I/O, computation, and sampling in large‑scale TGNN training, and demonstrate that existing system designs leave significant performance headroom.
\item \xin{We propose \textsc{FAST}, a holistic framework that co-designs system optimizations with TGNN semantics, including a SlimCache strategy that combines compression and caching for reduced host–device traffic, thread‑efficient graph operators that address load imbalance and low thread utilization in sparse temporal graphs (accelerating both aggregation and edgeSoftmax), and a topology‑aware CPU sampling strategy that leverages CPU core and cache hierarchies to improve locality and throughput.}
\item \xin{We evaluate \textsc{FAST} on real‑world large dynamic graphs and show that it achieves an average 2.1$\times$ speedup (up to 4.7$\times$) over state‑of‑the‑art systems without sacrificing model accuracy.}
\end{itemize}

%% file: text/background.tex
\section{\xin{Background}}
\subsection{Dynamic Graphs}

\xin{Dynamic graphs are a fundamental data structure for modeling timestamped interactions between entities in real‑world systems. A dynamic graph is typically represented as a sequence of timestamped events $\mathcal{G}_S = \{e_{t_1}, e_{t_2}, \cdots, e_{t_{|E|}}\}$, where each edge $e_t = (u, v)$ denotes an interaction between nodes $u$ and $v$ occurring at timestamp $t$ \citep{etc,simple,taser,swift_2025}. Based on the granularity of temporal modeling, dynamic graphs are broadly categorized into two types.}

\xin{\textbf{Continuous‑Time Dynamic Graphs (CTDGs)} treat each interaction as an independent event at a continuous time point, commonly expressed as $\alpha(t) = (u, v, \mathbf{e}_{uv}(t), t)$, where $\mathbf{e}_{uv}(t)$ is the edge feature vector at time $t$ \citep{tgn}. CTDGs naturally capture fine‑grained evolution patterns, including edge additions, deletions, updates, and node‑level events modeled as self‑loops. }

\xin{\textbf{Discrete‑Time Dynamic Graphs (DTDGs)} approximate the continuous process by partitioning the timeline into equal‑length time windows (or snapshots) \citep{Chen2023NeutronStreamAD}, converting each window into a static graph. While DTDGs are often computationally more efficient, they sacrifice some temporal continuity compared to CTDGs.}

\subsection{Temporal Graph Neural Networks}
\xin{TGNNs are designed to learn node representations from dynamic graphs in an end‑to‑end manner. TGN \citep{tgn} and TGAT \citep{tgat} are representative models for CTDGs, which process event streams through temporal encoding modules and generate dynamic node embeddings at arbitrary time points. DySAT \citep{sankar2020dysat} targets DTDGs, aggregating information both within each snapshot (structural dimension) and across snapshots (temporal dimension) to capture evolving structural patterns. TGL \citep{tgl} argues that DTDGs can be viewed as discretized CTDGs and designs a unified training framework that supports both CTDG‑ and DTDG‑based TGNNs.}

\yushu{These models compute node embeddings via temporal message passing. For a target node and its temporal neighbors, TGNNs perform iterative \emph{sample‑aggregate‑update} steps, where forward aggregation follows the GraphSAGE paradigm \citep{tgn, Lo2021EGraphSAGEAG}:}
\begin{equation}
h_v^{(l)} = \mathrm{AGG}\left(\left\{M\left(h_u^{(l-1)}, h_v^{(l-1)}, e_{uv}, t\right) \mid u \in \mathcal{N}_\mathrm{in}(v)\right\}\right).
\label{eq:forward_agg}
\end{equation}
\xin{where node $v$’s embedding at layer $l$ is updated by aggregating messages from its incoming neighbors $\mathcal{N}_\mathrm{in}(v)$, $M(\cdot)$ is the message function, $e_{uv}$ denotes edge features, and $t$ is the timestamp.}

\xin{This work focuses on TGNNs applied to CTDGs, where TGNNs operate on an event‑stream representation and rely on chronological order batch training for scalable learning. Similar to TGL \citep{tgl}, our framework can support TGNNs on DTDGs but its performance remains suboptimal. We leave the design of optimizations specifically for DTDG-style training as future work.}

%% file: text/motivation.tex
\section{Motivation}
\xin{Training TGNN on large dynamic graphs shows fundamental inefficiencies across the full training pipeline. While prior work has explored optimizations in individual stages, we find that performance remains bottlenecked by a combination of memory I/O overhead, irregular computation, and suboptimal sampling. In this section, we revisit these bottlenecks through measurement and analysis, and identifies key opportunities for a unified, system-level solution.}

\subsection{\xin{Rethinking Memory I/O}}
\label{sec:motivate_io}

\begin{figure}[t]
 \includegraphics[width=\columnwidth]{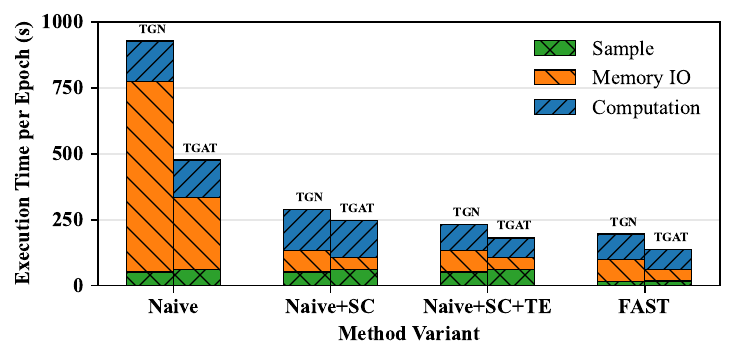}
 \caption{The execution time breakdown analysis on WIKITALK dataset of TGAT and TGN(generated from 5 epochs).}
 \label{fig:moti_break}
\end{figure}

\xin{We begin by analyzing the execution time breakdown of representative TGNN models (e.g., TGAT and TGN) on real-world datasets. As shown in Figure~\ref{fig:moti_break}, memory I/O dominates end-to-end training time, accounting for up to 78\% of total execution. This overhead stems from repeated host–device transfers of dynamically sampled subgraphs and associated features, which are fundamentally constrained by limited interconnect bandwidth (e.g., PCIe) \citep{Zhu2024FastGLAG}.}
\xin{Existing approaches attempt to reduce data movement via either caching or compression \citep{etc,simple,taser}. Caching-based methods exploit cross-batch reuse by storing frequently accessed data in GPU memory, while compression-based methods reduce within-batch redundancy by transmitting compact representations. Although effective in isolation, these strategies fail to fully utilize available GPU memory or exploit the full spectrum of redundancy present in TGNN workloads.}

\xin{In practice, large-scale TGNN training operates in a regime between two extremes: GPU memory is insufficient for full caching, yet not fully utilized during execution. This creates an opportunity to jointly leverage caching and compression under a unified design. Moreover, we observe that nodes and edges exhibit distinct reuse patterns during batched training. However, existing systems either treat them uniformly or prioritize node caching \citep{simple,taser}, leading to suboptimal cache utilization.}
\xin{To validate this observation, we perform a pre-sampling analysis to quantify redundancy patterns across batches. To characterize redundancy in sampled subgraphs, we adopt two metrics. First, we use the overlap ratio $M_{ij}$ \citep{Zhu2024FastGLAG} measuring the fraction of shared nodes (or edges) between two subgraphs $i$ and $j$. Second, we define the repetition rate
$R_i$ to capture the proportion of duplicate IDs within a subgraph. Formally, 
\[
M_{ij} = \frac{N_o}{\min(N_i, N_j)}, \qquad
R_i = 1 - \frac{N_u}{N_i}.
\]
where $N_o$ denotes the number of overlapping IDs, $N_u$ denotes the number of unique IDs in subgraph $i$, and $N_i$ and $N_j$ denote the total number of IDs in subgraphs $i$ and $j$, respectively.}

\xin{The results, shown in Table~\ref{tab:moti_degree_unblance}, present overlap and repetition statistics across four datasets. We observe that both $M$ and $R$ are consistently higher for nodes than for edges, which is consistent with prior observations based on Jaccard similarity in GNNFlow \citep{gnnflow}. This behavior arises because a single node typically participates in multiple interactions within a sampled subgraph, leading to higher redundancy compared to edges.}

\xin{These findings have direct system implications. A high repetition rate indicates greater potential for within-batch compression, while a high overlap ratio suggests improved effectiveness of cross-batch caching. Together, they highlight the need for a unified design that (1) integrates caching and compression to reduce host–device data movement, and (2) explicitly accounts for the heterogeneous reuse patterns of nodes and edges to maximize cache efficiency under limited memory budgets.}

\subsection{\xin{Inefficiency in Graph Computation}}
\label{sec:motivate_compute}

\begin{figure}[t]
 \includegraphics[width=\columnwidth]{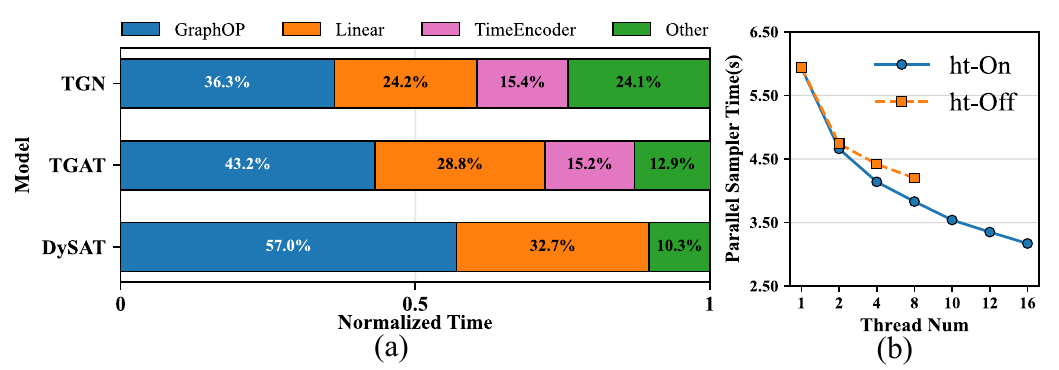}
 \caption{(a)~Forward pass time breakdown on WIKITALK. (b)~Sampler time on LASTFM with hyper-threading on/off.}
 \label{fig:moti_break_gop}
\end{figure}

\xin{After alleviating memory I/O overhead, computation becomes dominant, as shown in Figure~\ref{fig:moti_break}. A detailed operator-level breakdown in Figure~\ref{fig:moti_break_gop}(a) shows that graph operators such as aggregation(AGG) and edge-softmax(ESM) account for up to 57\% of total computation time.}
\xin{Unlike dense tensor operations, these operators exhibit irregular memory access patterns and highly skewed workloads due to the sparsity of graphs \yushu{\citep{fuseGNN}}. Existing frameworks rely on general abstractions (e.g., gSpMM/gSDDMM) \citep{dgl,GNNone}, which limit opportunities for operator-specific optimization. While prior work on static graphs improves performance through kernel fusion and memory optimization \yushu{\citep{dGNN,Zhu2024FastGLAG}}, these techniques do not generalize well to TGNNs with highly sparse and dynamic topologies.}

\begin{figure}[t]
 \includegraphics[width=\columnwidth]{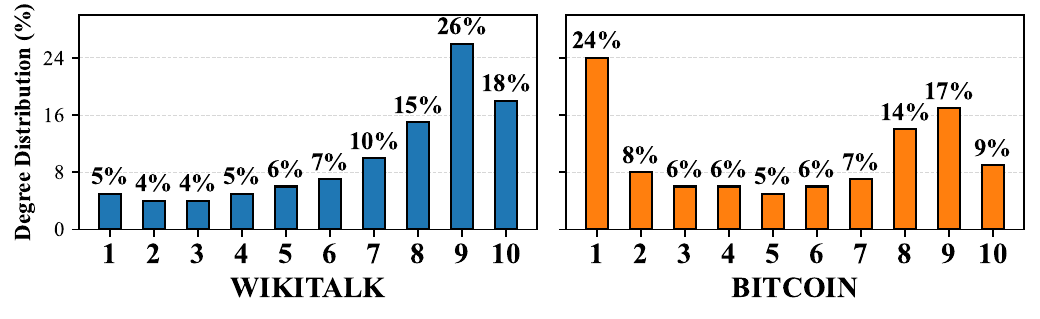}
 \caption{Distributions of root node degree.}
 \label{fig:degree_distribution}
\end{figure}

\xin{Our analysis reveals two key inefficiencies. First, node-centric parallelization leads to severe load imbalance due to skewed degree distributions. As illustrated in Figure~\ref{fig:degree_distribution}, dynamic graphs often exhibit long-tailed or small-degree distributions. We quantify this imbalance using the unbalance rate defined in Equation~(\ref{eq:unbalance_rate}). Let $R_d$ be the proportion of degree-$d$ nodes, and $D_{\max}$ be the maximum sampled degree.
The unbalance score for degree-$d$ nodes is $\mathrm{US}(d)=R_d(D_{\max}-d)$. The worst-case imbalance occurs when all nodes have degree 1, yielding $\mathrm{US}_{\mathrm{worst}}=D_{\max}-1$.
Thus, the unbalance rate is:
\begin{equation}\label{eq:unbalance_rate}
\mathrm{UR}(d)=\frac{\mathrm{US}(d)}{\mathrm{US}_{\mathrm{worst}}}
=\frac{R_d(D_{\max}-d)}{D_{\max}-1}.
\end{equation}
A higher $\mathrm{UR}(d)$ indicates more severe load imbalance. We report the results in Table~\ref{tab:moti_degree_unblance}. For example, the WIKITALK dataset reaches up to 35.6\% of the worst-case imbalance. This imbalance directly leads to low hardware utilization; Table~\ref{tab:kernel_ncu_naive} shows that the average active threads per warp for the aggregation operator is only 57.6\%, meaning that nearly half of the threads within a warp remain idle. Datasets with higher unbalance rates are expected to suffer from even more severe idling.}

\xin{Second, the small-degree nature of sampled temporal subgraphs results in low efficiency for warp-level primitives. For edge softmax, reduction operations are typically mapped to fixed-width warp units \citep{dGNN}. However, when node degrees are small, many threads remain idle. As shown in Table~\ref{tab:moti_degree_unblance}, the theoretical thread efficiency ranges only from 53\% to 76\% under typical settings. Although measured warp activity appears high (e.g., 78.5\% in Table~\ref{tab:kernel_ncu_naive}), a significant fraction of threads do not contribute to effective computation.\textbf{}}
\xin{These inefficiencies not only degrade compute utilization but also limit memory locality and reduce the number of active warps per SM. Collectively, they motivate the need for TGNN-specific thread-efficient operator designs that jointly optimize workload balance, thread utilization, and memory access efficiency.}

\begin{table}[htbp]
  \centering
  \caption{Match, repeat, imbalance rates and degree efficiency of graph datasets}
  \small
  \begin{tabular}{@{}l@{\extracolsep{\fill}}ccccccc@{}}
    \toprule
    \multirow{2}{*}{Dataset} & 
    \multicolumn{2}{c}{Match (\%)} & 
    \multicolumn{2}{c}{Repeat (\%)} & 
    \multirow{2}{*}{\makecell[c]{Unbalance \\ rate (\%)}} & 
    \multirow{2}{*}{\makecell[c]{Avg. degree\\(Thread efficiency)}} \\
    \cmidrule(lr){2-3} \cmidrule(lr){4-5}
    & node & edge & node & edge & & \\
    \midrule
    LASTFM   & 91.8 & 19.9 & 93.5 & 75.0 & 36.5  & 6.83(68\%)  \\
    WIKITALK & 71.5 & 41.3 & 91.0 & 73.0 & 35.6  & 7.56(76\%)  \\
    BITCOIN  & 64.1 & 38.3 & 84.0 & 64.0 & 52.0  & 5.34(53\%)  \\
    GDELT    & 94.6 & 47.4 & 96.5 & 76.0 & 42.3  & 7.08(70\%)  \\
    \bottomrule
  \end{tabular}
  \label{tab:moti_degree_unblance}
\end{table}

\begin{table}[htbp]
  \centering
  \caption{Performance of AGG and ESM on WIKITALK.}
  \small
  \begin{tabular}{@{}l@{\extracolsep{\fill}}cccccc@{}}
    \toprule
    Kernel\text{ } & L1 Cache & L2 Cache & Act. warps & Act. threads/warp \\
    \midrule
    Aggregation  & 80.45\% & 56.33\% & 39.76 / SM & 18.94 (57.6\%) \\
    Edge-Softmax  & 37.62\% & 73.81\% & 11.24 / SM & 25.11 (78.5\%) \\
    \bottomrule
  \end{tabular}
  \label{tab:kernel_ncu_naive}
\end{table}
%
\subsection{\xin{Opportunity in Sampling}}
\label{sec:motivate_sampler}

\xin{As memory I/O and computation are progressively optimized, the cost of temporal neighbor sampling becomes increasingly visible (Figure~\ref{fig:moti_break}). Existing CPU-based samplers (e.g., TGL) leverage multi-threading to exploit data parallelism \citep{tgl,etc,simple}, but do not fully utilize modern CPU cache hierarchies. Meanwhile, GPU-based samplers achieve high performance through specialized designs \citep{gnnflow,msPipe,taser}, but often sacrifice generality and reusability.}
\xin{We empirically evaluate the impact of hardware features such as Hyper-Threading (HT) on sampling performance. As shown in Figure~\ref{fig:moti_break_gop}(b), enabling HT provides only marginal improvements, indicating that naive thread scaling is insufficient to fully exploit hardware capabilities\citep{ht_icpp}.}

\xin{We further observe that sampled subgraphs exhibit structural locality, where different root nodes share overlapping neighborhoods. Prior work such as NextDoor \citep{nextDoor_Jangda2020AcceleratingGS} exploits this property on GPUs using software-managed memory. However, such mechanisms are not directly available on CPUs.}
\xin{Instead, we identify an alternative opportunity: hardware-aware thread scheduling. Modern CPUs expose shared cache hierarchies across cores (e.g., L2/L3 caches), suggesting that carefully mapping sampling tasks to cores can improve cache locality. The key challenge is bridging the gap between abstract graph topology and physical hardware organization.}
\xin{To address this, we propose a topology-aware sampling strategy that analyzes overlap patterns in sampled subgraphs (via pre-sampling) and maps threads to CPU cores accordingly. By aligning computation with cache sharing domains, this approach improves cache hit rates and reduces memory access latency, enabling efficient and generalizable sampling acceleration.}

%% file: text/design.tex
\section{Design}

\subsection{Overview}
\label{sec:design_overview}

\begin{figure}
  \includegraphics[width=\columnwidth]{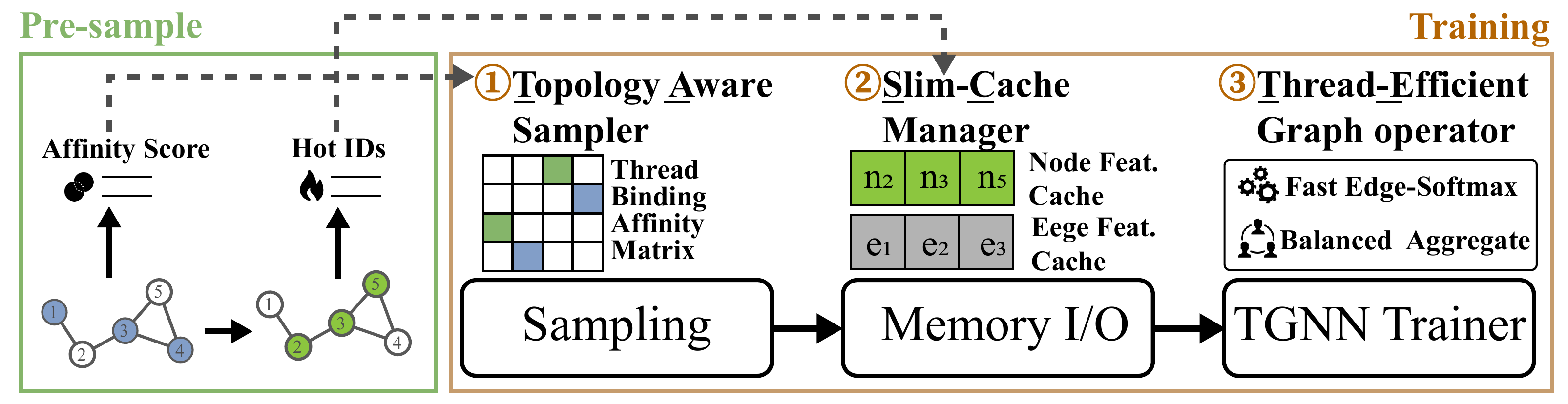}
  \caption{Overall architecture of FAST.}
  \label{fig:overview}
\end{figure}

\xin{We present FAST, a holistic framework that accelerates TGNN training by jointly addressing bottlenecks in sampling, memory I/O, and computation. Figure~\ref{fig:overview} illustrates the overall architecture. \yushu{FAST begins with a lightweight pre-sampling phase, which extracts two key pieces of information based on the target batch size: (1) thread affinity score for topology-aware sampler, and (2) redundancy statistics used to identify frequently accessed node and edge IDs (i.e., hot IDs).  Prior to training, the topology-aware sampler maps logical threads to physical CPU cores according to the affinity matrix, improving cache locality and reducing memory access latency. Simultaneously, the SlimCache manager selects frequently accessed node and edge features for GPU caching under a fixed memory budget using a greedy strategy based on hot IDs.}}

\yushu{During training, FAST first executes the pre-configured sampler. For memory I/O, SlimCache reuses cached data across batches, while uncached data are transferred in a compressed format to reduce host-device data movement and accelerate memory I/O. In the GPU trainer, }FAST employs thread-efficient graph operators tailored to the sparse and irregular structure of temporal graphs, improving workload balance and cache utilization through customized CUDA kernels. The above pipeline is executed iteratively over mini-batches until the end of each epoch. FAST is designed for a single-machine, single-GPU setting, providing an integrated solution for efficient large-scale TGNN training.

\subsection{\xin{SlimCache: Joint Caching and Compression for Efficient Memory I/O}}
\label{sec:design_sc}

\xin{Existing memory I/O optimization techniques \citep{etc,simple} treat caching and compression as independent strategies and overlook the distinct redundancy characteristics of nodes and edges in batched TGNN training, leading to suboptimal cache utilization. To address this limitation, we propose \emph{SlimCache}, a unified design that jointly exploits \textit{within-batch repetition} and \textit{cross-batch overlap} to minimize host-device data movement.}

\begin{figure}
  \includegraphics[width=\columnwidth]{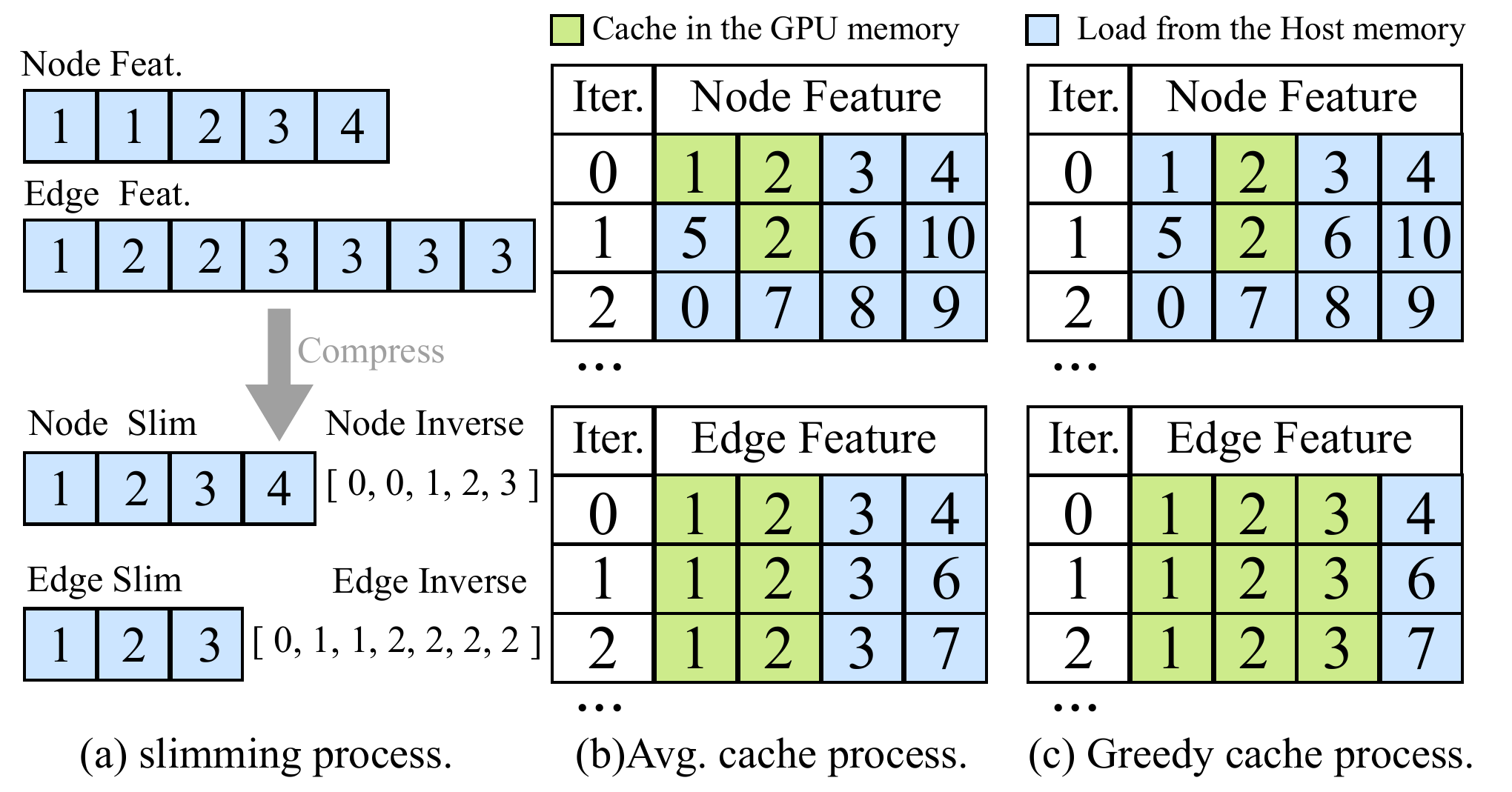}
  \caption{Process of slimming and cache.}
  \label{fig:design_greedy_process}
\end{figure}

\xin{SlimCache operates in two stages. First, during runtime, it applies ID-based compression to eliminate redundant data transfers within each batch. As illustrated in Figure~\ref{fig:design_greedy_process}(a), for $D$-dimensional node and edge features, only unique IDs and their corresponding features are transferred, along with an inverse index for reconstruction on the GPU. This reduces memory traffic from $12D$ to $7D$ in the example. Second, SlimCache leverages a GPU-resident cache to reuse frequently accessed data across batches. A naive strategy that equally partitions cache space between nodes and edges (Figure~\ref{fig:design_greedy_process}(b)) yields limited benefit (e.g., $9D$ cache hits over three iterations). However, as shown in Table~\ref{tab:moti_degree_unblance}, nodes and edges exhibit different redundancy patterns, suggesting that \yushu{rigid} allocation policies are suboptimal.}

\xin{To maximize cache effectiveness, we design a greedy selection strategy that determines cache placement based on access frequency. Specifically, we perform pre-sampling \citep{Yang2022GNNLabAF} over multiple mini-batches to collect access statistics for node and edge IDs. As described in Algorithm~\ref{alg:greedy_cache}, we first initialize access counters for nodes and edges (line 1), then accumulate their access frequencies across sampled batches (lines 2-4). The IDs are sorted in descending order of access frequency to obtain ranked lists (line 6). Given a fixed cache budget, we iteratively allocate space to nodes or edges based on their marginal gain in cache hits, \yushu{obtaining} the corresponding hot ID sets (lines 8-10). This process completes before training and determines the final cache layout.}

\xin{Figure~\ref{fig:design_greedy_process}(c) illustrates the behavior of the greedy strategy. By capturing higher cross-batch overlap in edges, SlimCache allocates more cache budget to edges, achieving higher overall cache efficiency. For example, the total feature hit count reaches $11D$, outperforming the equal-allocation strategy in Figure~\ref{fig:design_greedy_process}(b). More broadly, the redundancy analysis in Table~\ref{tab:moti_degree_unblance} shows that nodes and edges differ significantly in both repetition and overlap, reinforcing the need for differentiated treatment.}
\xin{By tightly integrating compression and caching, SlimCache simultaneously exploits within-batch redundancy and cross-batch reuse. This unified design substantially reduces memory I/O traffic and improves data access efficiency, enabling scalable TGNN training on large dynamic graphs.}

\begin{algorithm}[t]
\caption{Greedy selection strategy}
\label{alg:greedy_cache}
\small
\begin{algorithmic}[1]
\Require Cache budget $\alpha$, batches $\{(\mathcal{N}_k, \mathcal{E}_k)\}_{k=1}^b$, node set $V$, edge set $E$
\Ensure Hot node set $\mathcal{H}_N$, hot edge set $\mathcal{H}_E$
\State $\mathbf{f}_v \gets \mathbf{0}_{|V|}, \quad \mathbf{f}_e \gets \mathbf{0}_{|E|}$
\For{$(\mathcal{N}_k, \mathcal{E}_k) \in$ batches} \Comment{Count frequencies}
    \State $\mathbf{f}_v[\mathcal{N}_k] \gets \mathbf{f}_v[\mathcal{N}_k] + 1$
    \State $\mathbf{f}_e[\mathcal{E}_k] \gets \mathbf{f}_e[\mathcal{E}_k] + 1$
\EndFor
\State $(\mathbf{s}_v, \mathbf{s}_e) \gets (\text{argsort}(\mathbf{f}_v), \text{argsort}(\mathbf{f}_e))$
\State $(i, j) \gets (0, 0)$
\While{$i + j < \alpha(|V| + |E|)$} \Comment{Greedy selection under budget}
    \If{$\mathbf{f}_v[\mathbf{s}_v[i]] > \mathbf{f}_e[\mathbf{s}_e[j]]$}
        \State $\mathcal{H}_N \gets \mathcal{H}_N \cup \{\mathbf{s}_v[i]\}$
        \State $i \gets i + 1$
    \Else
        \State $\mathcal{H}_E \gets \mathcal{H}_E \cup \{\mathbf{s}_e[j]\}$
        \State $j \gets j + 1$
    \EndIf
\EndWhile
\State \textbf{return} $\mathcal{H}_N, \mathcal{H}_E$
\end{algorithmic}
\end{algorithm}

\subsection{\xin{Thread-Efficient Graph Operators}}
\label{sec:design_compute}

\xin{As discussed in Section~\ref{sec:motivate_compute}, the computation stage of TGNN training is dominated by thread-inefficient graph operators. In particular, the aggregation(AGG) operator suffers from severe load imbalance, which reduces active threads per warp, while the edge-softmax(ESM) operator incurs low thread efficiency during reduction because of the small-degree property of sampled temporal subgraphs. These issues jointly lead to poor SM utilization and limit the performance of the computation stage. To address them, we redesign the two critical operators with TGNN-specific execution strategies: an edge-centric AGG scheme to eliminate load imbalance, and a thread-efficient reduction scheme for \textsc{ESM} to improve thread utilization and expose more locality for the GPU cache.}

\xin{\textbf{Balanced Aggregate.} \yushu{TGNN aggregation follows the GraphSAGE paradigm by summing neighbor embeddings for each target node~\cite{tgl}, as defined in Equation ~\eqref{eq:forward_agg}.} Unlike conventional GNNs, TGNNs incorporate temporal and edge information during message passing \citep{tgat,tgn,sankar2020dysat}, and the corresponding message function for \textsc{AGG} can be simplified to directly copying the neighbor embedding, i.e., the \texttt{copy\_u} primitive in DGL. Therefore, the AGG operator in TGNNs reduces to a sum over neighbor embeddings, where for a target node $v$ we have $h_v = \sum_{u \in \mathcal{N}_\mathrm{in}(v)} v_u$
with $\mathcal{N}_\mathrm{in}(v)$ denoting the set of incoming neighbors of node $v$.}

\begin{figure}
  \includegraphics[width=\columnwidth]{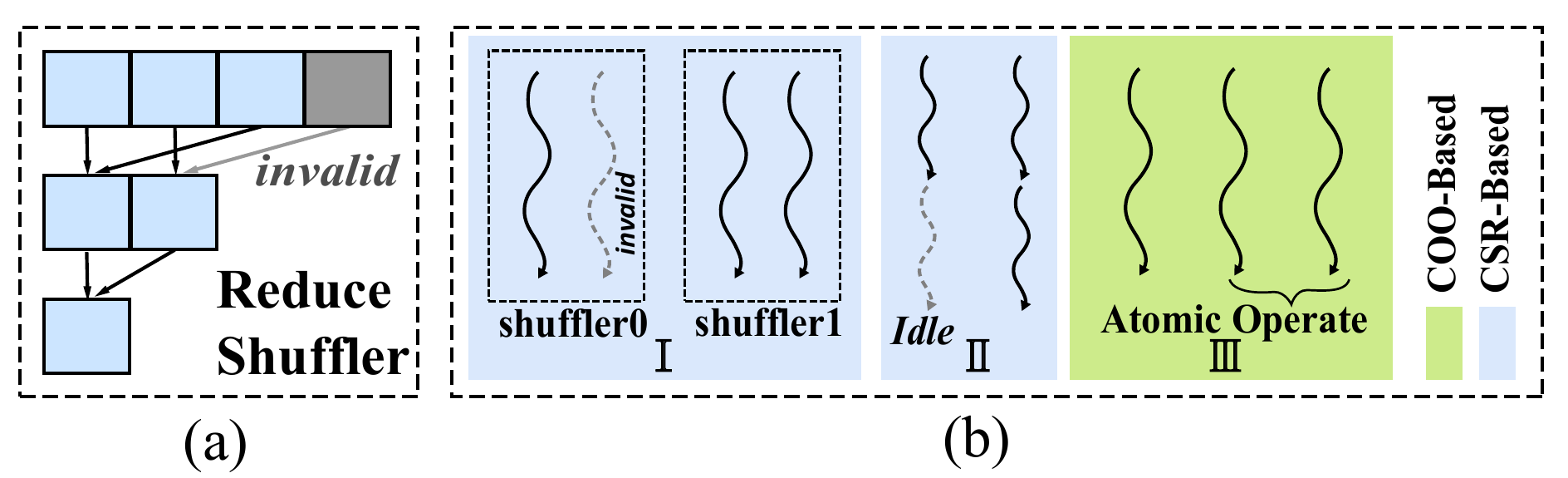}
  \caption{(a) Invalid thread in shuffler.(b) Reduction using COO(Coordinate) or CSR(Compressed Sparse Row) formats.}
  \label{fig:agg_esm}
\end{figure}

\xin{Existing optimizations such as FastGL\citep{Zhu2024FastGLAG} improve \textsc{AGG} mainly at the memory-access level by caching frequently used features in shared memory(SMEM). However, for TGNN workloads, the feature dimension limits how many nodes can be processed within one block; for example, when the feature dimension is $D=128$, at most 8 nodes can be processed per block. As a result, increasing cache hit rate through a larger per-block working set is difficult. We therefore focus on the computation-level bottleneck. In the conventional node-parallel implementation, each thread iterates over the neighbors of one node, as shown in Figure~\ref{fig:agg_esm}(b)-\uppercase\expandafter{\romannumeral2}. Because neighborhood sizes vary substantially, threads finish at different times and many become idle, leading to the load imbalance described in Section~\ref{sec:motivate_compute}. Although a CSR-based balancing strategy could assign additional neighborhoods to idle threads, such fine-grained scheduling is ill-suited to TGNNs, where sampled subgraphs are sparse and the overhead of task redistribution would offset the benefit.}

\xin{To eliminate this imbalance, we adopt an edge-centric execution scheme based on the COO representation, as shown in Figure~\ref{fig:agg_esm}(b)-\uppercase\expandafter{\romannumeral3}. In this design, each thread handles one edge independently, which distributes work more evenly across threads and removes the degree-dependent imbalance of the node-parallel method. The main concern is the need to combine partial results from multiple edges into the same destination node, which requires atomic operations such as \texttt{atomicAdd}. However, for TGNNs, this contention is manageable. For a graph $G(V,E)$ with $n$ nodes, let $d_{\text{in}}(v)$ denote the in-degree of node $v$. The atomic blocking delay is bounded by
\[
T_{\text{atomic-block}} \leq \max_{v \in V} \left(d_{\text{in}}(v)-1\right) = d_{\text{max}} - 1.
\]
where $d_{\text{max}}=\max_{v \in V} d_{\text{in}}(v)$. This indicates that atomic contention scales linearly with the maximum degree. Since dynamic graph workloads are typically small-degree, the blocking overhead remains limited. As shown in Table~\ref{tab:moti_degree_unblance}, the maximum average degree across training batches is 7.56, and the maximum degree in a batch does not exceed the sampling bound of 10. Under such conditions, the contention cost of the COO-based design is negligible in practice. More importantly, because \textsc{AGG} in TGNNs is additive, each thread performs only a single atomic update to global memory(GMEM), avoiding the need for complex SMEM buffering. This makes the edge-centric design both simple and efficient.}

\xin{\textbf{Edge Softmax.} \textsc{ESM} computes a numerically stable softmax over the edges of each sampled subgraph \citep{milakov2018online_safe}, involving a maximum reduction, a sum reduction, and an element-wise division. In the existing dGNN\citep{dGNN} implementation, the reduction is performed with warp shuffle, as illustrated in Figure~\ref{fig:agg_esm}(b)-\uppercase\expandafter{\romannumeral1}. The reduction width must be set according to the maximum neighbor count, and threads assigned to positions beyond the actual degree remain active but perform ineffective work.}

\xin{A natural alternative is to reuse the COO-style edge-parallel reduction used for \textsc{AGG}. However, this is not suitable for \textsc{ESM}. First, implementing the three stages of safe softmax with atomic reads and writes to GMEM would introduce substantial memory-access overhead. Second, using SMEM to optimize repeated GMEM accesses would require additional preprocessing to ensure that each node's neighborhood is not split across blocks, which adds nontrivial overhead. Therefore, the COO-based design does not provide a good tradeoff for \textsc{ESM}.}
\xin{Instead, we refine the CSR-style thread partitioning strategy used by FastGL. As shown in Figure~\ref{fig:agg_esm}(b)-\uppercase\expandafter{\romannumeral2}, each thread iterates over multiple reduction elements, which removes invalid active threads and improves thread utilization. This design is particularly effective for TGNNs because the thread idling introduced by small degrees is much less severe in \textsc{ESM} than \textsc{AGG}. Let $|E|$ be the number of edges, $D_f$ the feature dimension, and $H$ the number of attention heads. For the same set of edges, the ratio of idle threads between \textsc{ESM} and \textsc{AGG} is approximately $\mathrm{IdleNum}_{\mathrm{ESM}} / \mathrm{IdleNum}_{\mathrm{AGG}} = H / D_f \ll 1$. For typical settings such as $H=2$ and $D_f=128$, the number of idle threads in \textsc{ESM} is less than 2\% of that in \textsc{AGG}. In addition, this thread-loop design also increases the per-block node count. Given a block supporting up to 1024 threads, the original shuffle reduction processes 64 nodes per block (due to a 16-element shuffler), whereas the thread-loop version processes 1024 nodes, achieving a 16$\times$ increase.Combined with the within-batch repetition observed in Section~\ref{sec:motivate_io}, this larger working set exposes more spatial locality among sampled subgraphs and improves cache hit rates.}

\xin{Overall, our computation design tailors the \textsc{AGG} and \textsc{ESM} operators to the structural properties of TGNN workloads. The balanced \textsc{AGG} eliminates load imbalance through edge-centric execution, while the thread-efficient \textsc{ESM} improves reduction efficiency and enlarges the per-block working set. Together, these optimizations reduce both computation and memory-access overheads in forward and backward propagation, enabling efficient TGNN training on sparse dynamic graphs.}

\subsection{\xin{Topology-Aware Sampling}}
\label{sec:design_sample}

\xin{As discussed in Section~\ref{sec:motivate_sampler}, accelerating the sampling stage requires mapping logical sampling threads to physical CPU cores in a topology-aware manner, so as to reduce memory access latency by improving CPU cache locality. Unlike GPUs, CPU cache resources cannot be explicitly controlled. However, the placement of sampling tasks over physical cores can be managed. Modern processors support fine-grained thread binding through OpenMP 4.0 \citep{icpp_openmp}, which provides the basic mechanism for such control.}

\begin{figure}
  \includegraphics[width=\columnwidth]{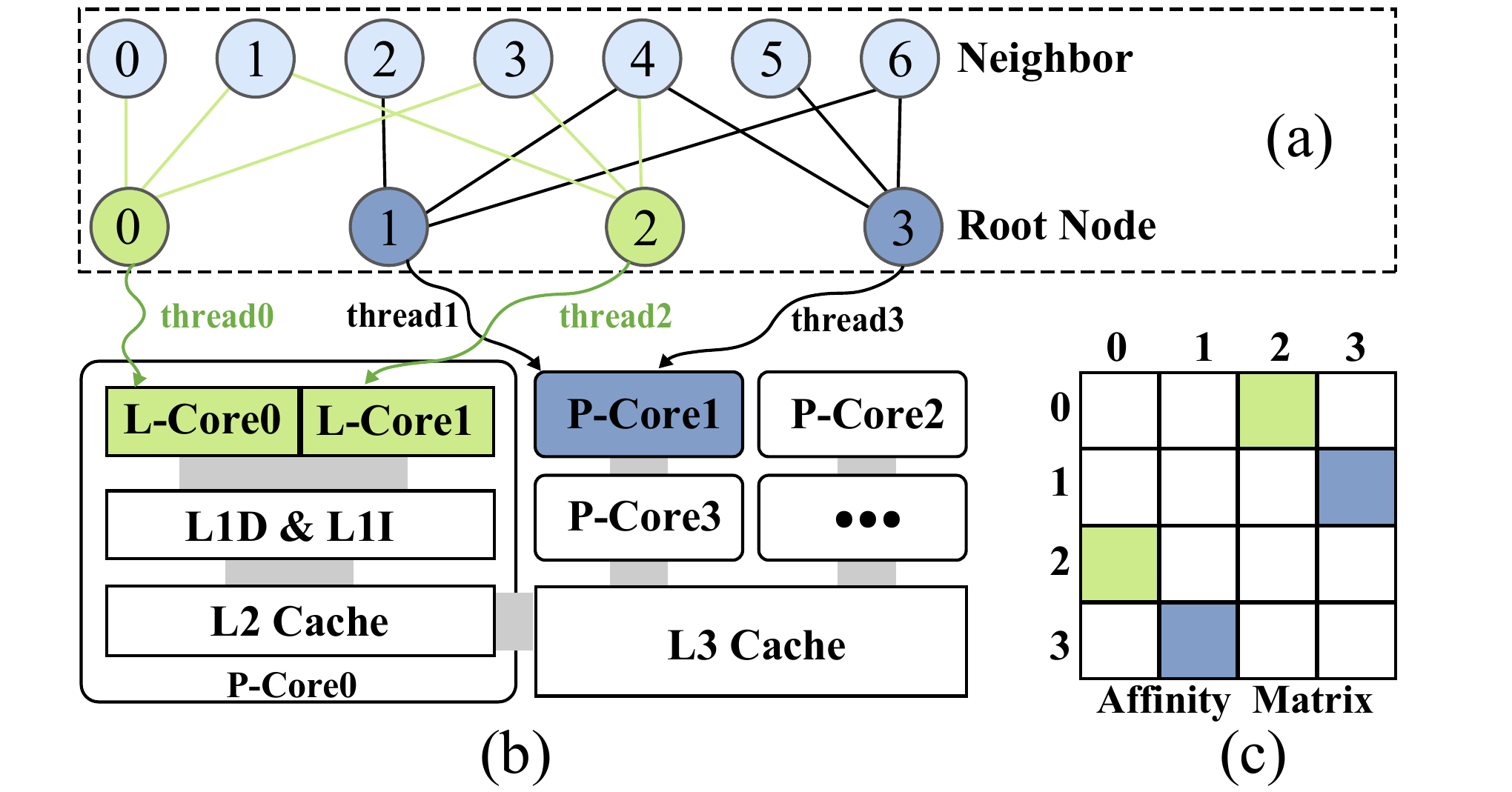}
  \caption{Topology-aware binding and affinity matrix.}
  \label{fig:cpu}
\end{figure}

\xin{A key challenge is that batched TGNN training executes sampling repeatedly across many mini-batches within an epoch. Recomputing thread bindings at runtime would introduce unnecessary overhead and offset the performance gains from improved cache hit rates. We therefore determine the binding configuration once before training begins and reuse it throughout the epoch. The remaining question is how to characterize affinity among sampling threads based on the dynamic topology of all training batches. To this end, we design a topology-aware strategy that analyzes the overlap patterns among the subgraphs processed by different threads. As illustrated in Figure~\ref{fig:cpu}(a), each root node and its sampled neighborhood are assigned to a specific logical thread. Figure~\ref{fig:cpu}(b) shows the cache organization of a hyper-threaded CPU, where each performance core (P-Core) contains two logical cores (L-Core0 and L-Core1). To exploit the shared L2 cache, threads that process highly similar sampled subgraphs should be bound to the same P-Core, such as threads 0 and 2 highlighted in green. To capture this relationship, we construct the thread affinity matrix shown in Figure~\ref{fig:cpu}(c), where rows and columns correspond to thread IDs and each entry denotes the similarity between the sampled subgraphs processed by the corresponding pair of threads. Based on this matrix, we perform pairwise matching to produce the final logical-core binding list. Specifically, we use the Blossom algorithm \citep{blossom_algo_shoemaker2016edmonds} to find the maximum-weight matching that maximizes the total affinity across all thread pairs.}

\xin{Algorithm~\ref{alg:topology_aware} summarizes the construction of the thread affinity matrix for batched training. We first initialize the affinity matrix according to the number of threads (line 1). We then iterate over the sampling root nodes of $R$ batches. For each batch, every thread processes a contiguous chunk of nodes, following OpenMP static scheduling (lines 2-4). Given the chunk size, we compute the starting position for each thread within the batch (lines 5-8). Next, we traverse the upper triangular part of the matrix to compute the affinity between each pair of threads and write the result into the matrix (lines 9-13). The lower triangular part is filled to enforce symmetry, and the diagonal entries are set to -1 to prevent self-pairing (lines 14-15). Finally, the Blossom algorithm \citep{blossom_algo_shoemaker2016edmonds} generates the binding list used for thread-to-core placement (line 16).}

\begin{algorithm}[t]
\caption{Topology-aware thread binding}
\label{alg:topology_aware}
\small
\begin{algorithmic}[1]
\Require Root node IDs for all training batches: $\mathcal{R}$, number of threads: $T$, 
sub-graph similarity function: $\text{simFunc}(u,v,\text{nodes})$, blossom algorithm: $\text{BlossomSelect}(\mathbf{M})$
\Ensure Thread binding list: $\mathcal{B}$
\State Initialize affinity matrix $\mathbf{A} \gets \mathbf{0}^{T \times T}$
\For{each batch $b = 1$ to $|\mathcal{R}|$}
    \State $\mathbf{V}_b \gets \mathcal{R}[b]$ \Comment{Nodes in batch $b$}
    \State $C \gets \lfloor |\mathbf{V}_b| / T \rfloor$ \Comment{Chunk size per thread}
    \For{$k = 0$ to $C-1$}
        \For{$i = 0$ to $T-1$}
            \State $u_i \gets i \cdot C + k$
        \EndFor
        \ForAll{unordered pairs $(i,j)$ where $0 \leq i < j < T$}
            \State $\mathbf{A}[i][j] \gets \mathbf{A}[i][j] + \text{simFunc}(u_i, u_j, \mathbf{V}_b)$
        \EndFor
    \EndFor
\EndFor
\State Symmetrize $\mathbf{A}$: $\mathbf{A}[j][i] \gets \mathbf{A}[i][j]$ for all $0 \leq i < j < T$
\State Set diagonal: $\mathbf{A}[i][i] \gets -1$ for $i = 0$ to $T-1$
\State $\mathcal{B} \gets \text{BlossomSelect}(\mathbf{A})$
\State \Return $\mathcal{B}$
\end{algorithmic}
\end{algorithm}

\yushu{The \texttt{simFunc} (line~10) computes Jaccard similarity between $K$-hop neighborhoods in $O(S^K)$ time, where $S$ denotes the average number of neighbors. The affinity matrix construction costs $O(T^2 \cdot R \cdot C \cdot S^K)$ for $R$ batches, $C$ nodes per thread, and $T$ threads, while Blossom adds $O(T^3)$. The total complexity $O\big(T^2 \cdot R \cdot C \cdot S^K\big)$ (since $R \cdot C \cdot S^K \gg T$) is practically acceptable for typical parameters ($T=40$, $R=469$, $C=100$, $S=10$, $K=1$).
In our implementation, root nodes alone capture repetition patterns adequately. We set the pre-sampling hop number to $0$ (using only roots), reducing overhead and $K$-hop dependence. This lightweight strategy suffices for capturing thread affinity patterns, as demonstrated in our evaluation.}

%% file: text/impl.tex
\section{\xin{Implementation}}
\label{sec:impl_ct}
\yushu{\textsc{FAST} is built on the modular architecture of TGL\citep{tgl}, which uses DGL\citep{dgl} as the graph message-passing backend and provides dynamic graph learning support.To accelerate both I/O and computation, we develop a compression engine that generates compact triples(including \texttt{uni\_ID}, \texttt{inv\_Idx}, and CSR-style \texttt{ind\_ptr}) for SlimCache and CSR-Based graph operator execution.\footnote{Corresponding results will be shown in Section 6.3.}
We implement a Python class \texttt{cacheConfig()} to manage SlimCache’s greedy caching and compression logic. When GPU memory is insufficient, the feature cache ratio is scaled down according to the remaining space. Additionally, we design custom CUDA kernels for thread-efficient graph operators and expose them via user-facing forward and backward APIs for constructing different TGNN models. Through these modular APIs, each design component of \textsc{FAST} can be readily integrated into other TGNN training frameworks.}

%% file: text/evaluation.tex
\begin{table}[htbp]
\centering
\caption{Dataset statistic of the dynamic graphs. ($\mathcal{D}_v$ and $\mathcal{D}_e$ denote the dimensions of node features and edge features)}
\label{tab:dynamic_graphs}
\resizebox{0.48\textwidth}{!}{%
\begin{tabular}{lccccccc}
\toprule
\multirow{2}{*}{\makecell[c]{Graph}} & 
\multirow{2}{*}{\makecell[c]{$|V|$}} & 
\multirow{2}{*}{\makecell[c]{$|E|$}} & 
\multirow{2}{*}{\makecell[c]{$\mathcal{D}_v$}} & 
\multirow{2}{*}{\makecell[c]{$\mathcal{D}_e$}} & 
\multicolumn{3}{c}{\makecell[c]{Feature size (GB)}} \\
\cmidrule(lr){6-8}
& & & & & 
\makecell[c]{Vertex} & 
\makecell[c]{Edge} & 
\makecell[c]{Total} \\
\midrule
LastFM & 2K & 1.3M & 128 & 128 & 0.007 & 0.48 & 0.49 \\
WIKITALK & 1.1M & 7.8M & 172 & 172 & 0.73 & 5.01 & 5.74 \\
BITCOIN & 24.5M & 122.9M & 172 & 172 & 15.7 & 78.7 & 94.5 \\
GDELT & 17K & 191.3M & 413 & 182 & 0.02 & 129.7 & 130 \\
\bottomrule
\end{tabular}
}
\end{table}

\begin{table*}[t]
\centering
\caption{Comparison results of T-GNN training frameworks. Time refers to per-epoch execution time (s). The best average precision (\%) and the fastest execution time are marked in bold. "OOM" indicates out of memory.}
\label{tab:exp_pref_all}
\fontsize{3.0}{5}\selectfont 
\setlength{\tabcolsep}{4pt}  
\renewcommand{\arraystretch}{0.70}  
\setlength{\arrayrulewidth}{0.3pt}  
\resizebox{\textwidth}{!}{%
\begin{tabular}{l|l|lll|l|l|lll}
\hline
Dataset & Model & Framework & Time(s) & AP(\%) & Dataset & Model & Framework & Time(s) & AP(\%) \\
\hline
\multirow{10}{*}{LASTFM} & \multirow{4}{*}{TGN} & TGL & 153.17 (2.1×) & 86.25 & \multirow{10}{*}{BITCOIN} & \multirow{4}{*}{TGN} & TGL & 6824.28 (2.5×) & 90.42 \\
& & ETC & 94.79 (1.3×) & \textbf{86.47} & & & ETC & 4029.77 (1.4×) & 90.37 \\
& & SIMPLE & 94.85 (1.3×) & 86.13 & & & SIMPLE & OOM & N/A \\
& & FAST & \textbf{74.59} & 86.26 & & & FAST & \textbf{2784.33} & \textbf{90.50} \\
\cline{2-5} \cline{7-10}
 & \multirow{4}{*}{TGAT} & TGL & 77.85 (1.5×) & 86.73 & \multirow{4}{*}{} & \multirow{4}{*}{TGAT} & TGL & 2902.93 (2.4×) & \textbf{87.49} \\
& & ETC & 73.68 (1.4×) & 86.57 & & & ETC & 1411.99 (1.2×) & 87.35 \\
& & SIMPLE & 65.18 (1.3×) & 86.63 & & & SIMPLE & 3206.78 (2.6×) & 87.39 \\
& & FAST & \textbf{51.77} & \textbf{86.76} & & & FAST & \textbf{1217.96} & 87.44 \\
\cline{2-5} \cline{7-10}
 & \multirow{2}{*}{DySat} & TGL & 75.94 (1.5×) & \textbf{76.49} & \multirow{2}{*}{} & \multirow{2}{*}{DySat} & TGL & 8286.44 (1.4×) & 78.95 \\
& & FAST & \textbf{49.38} & 76.47 & & & FAST & \textbf{5921.24} & \textbf{78.91} \\
\hline
\multirow{10}{*}{WIKITALK} & \multirow{4}{*}{TGN} & TGL & 928.13 (4.7×) & 95.17 & \multirow{10}{*}{GDELT} & \multirow{4}{*}{TGN} & TGL & 31169.29 (4.2×) & 98.27 \\
& & ETC & 221.21 (1.1×) & 95.03 & & & ETC & OOM & N/A \\
& & SIMPLE & 237.44 (1.2×) & 94.98 & & & SIMPLE & OOM & N/A \\
& & FAST & \textbf{196.00} & \textbf{95.19} & & & FAST & \textbf{7385.34} & \textbf{98.32} \\
\cline{2-5} \cline{7-10}
 & \multirow{4}{*}{TGAT} & TGL & 476.99 (3.4×) & 90.84 & \multirow{4}{*}{} & \multirow{4}{*}{TGAT} & TGL & 16656.16 (3.2×) & 98.79 \\
& & ETC & 178.29 (1.3×) & \textbf{90.87} & & & ETC & OOM & N/A \\
& & SIMPLE & 215.26 (1.6×) & 90.74 & & & SIMPLE & OOM & N/A \\
& & FAST & \textbf{138.86} & 90.79 & & & FAST & \textbf{5213.77} & \textbf{98.82} \\
\cline{2-5} \cline{7-10}
 & \multirow{2}{*}{DySat} & TGL & 581.64 (1.8×) & 88.28 & \multirow{2}{*}{} & \multirow{2}{*}{DySat} & TGL & 35044.56 (2.3×) & 98.66 \\
& & FAST & \textbf{314.62} & \textbf{88.31} & & & FAST & \textbf{15056.18} & \textbf{98.68} \\
\hline
\end{tabular}%
}
\end{table*}

\section{Evaluation}
\subsection{Experimental Setup}

\textbf{Environments.}
\xin{ We conduct all experiments on a server equipped with dual Intel Xeon Gold 6133 CPUs (2x40 cores in total) running at 2.50 GHz, 512 GB of DRAM, and a single NVIDIA A100 GPU with 40 GB of GDDR6 VRAM. The software stack includes Python 3.8, PyTorch 2.1.2, DGL 0.9.1, and CUDA 11.8. Each reported result is averaged over five independent runs.}

\textbf{Datasets and Models.} 
\xin{As summarized in Table~\ref{tab:dynamic_graphs}, we evaluate \textsc{FAST} on four large-scale temporal graph datasets spanning different domains and graph characteristics. \textit{LastFM} \citep{lastfm_kumar2019predicting} captures listener--music interactions over a month. \textit{Wiki-Talk} \citep{ wiki2_paranjape2017motifs} records interactions among Wikipedia users on talk pages. \textit{Bitcoin} \citep{bitcoin1_kondor2014rich, bitcoin2_rossi2015network} is a subset of the Bitcoin transaction network. \textit{GDELT} is a near-billion-scale temporal knowledge graph derived from GDELT 2.0 \citep{gdelt_leetaru2013gdelt}, which models large-scale global events and interactions.}

\xin{We evaluate \textsc{FAST} using three representative TGNN backbones with different computational patterns: TGAT \citep{tgat}, which encodes temporal information with random Fourier features and attention-based aggregation; TGN \citep{tgn}, which maintains a memory vector for each node to model temporal evolution; and DySAT \citep{sankar2020dysat}, which applies self-attention over both structural neighborhoods and temporal dynamics.}

\textbf{Baselines and Settings.}
\label{sec:exp_config}
\xin{To assess the effectiveness of \textsc{FAST}, we compare against three state-of-the-art TGNN training frameworks: TGL \citep{tgl}, ETC \citep{etc}, and SIMPLE \citep{simple}. We exclude SWIFT \citep{swift_2025} from the main performance comparison because its performance is primarily constrained by disk bandwidth; we include it only in the overhead analysis in Section~\ref{sec:exp_overhead} to ensure a fair comparison. For component-wise analysis, we additionally use individual optimizations from other systems, including the caching design of TASER \citep{taser} for I/O comparison, as well as graph operator optimizations from static graph systems such as dGNN \citep{dGNN} and FastGL \citep{Zhu2024FastGLAG}.}

\xin{To ensure a fair comparison, we evaluate \textsc{FAST} and all baselines under the same training settings. For all TGNN models, we adopt top-$k$ recent neighbor sampling \citep{tgn} with $k=10$. We set the number of sampling threads to 8, following the default configurations of the baselines. All models use a standard 2-layer message-passing architecture \citep{tgn,tgat,sankar2020dysat}, with a batch size of 2000. We focus on the link prediction task \citep{etc}, and report test-set Average Precision (AP) as the primary accuracy metric. Each model is trained for 10 epochs, consistent with prior studies \citep{tgl,etc,simple, swift_2025}. Framework-specific hyperparameters are set to the recommended defaults of each baseline. For example, the cache budget ratio in SIMPLE, defined as the fraction of total input data that can be retained in memory, is set to its default value of 0.1.}

\subsection{Overall Performance}
\label{sec:exp_overall}

\xin{\textbf{Training Efficiency and Accuracy.} We first compare \textsc{FAST} against the baseline frameworks on end-to-end TGNN training. Table~\ref{tab:exp_pref_all} summarizes the results across different datasets and models. In terms of model quality, \textsc{FAST} achieves average precision (AP) values similar to those of all baseline systems, indicating that the introduced optimizations do not compromise model accuracy. In terms of training efficiency, however, \textsc{FAST} consistently outperforms the baselines across all three models, achieving an average speedup of \yushu{2.6$\times$ (up to 4.7$\times$)}  over TGL, \yushu{1.3$\times$ (up to 1.4$\times$)} over ETC, and \yushu{1.6$\times$ (up to 2.6$\times$)} over SIMPLE. \yushu{These gains are enabled by the combined effect of our SlimCache, thread-efficient operators, and topology-aware sampling, which together reduce memory I/O, computation overhead, and sampling latency.}}

\xin{On larger graphs with more than 100 million edges, \textsc{FAST} continues to deliver strong speedups. For the TGN model on BITCOIN and GDELT, \textsc{FAST} achieves \yushu{2.5$\times$} and \yushu{4.2$\times$} speedup on average over TGL, respectively. In contrast, ETC and SIMPLE encounter significant overhead that degrades their practical performance. For example, ETC runs into an Out-of-Memory (OOM) condition on GDELT, as its pipeline parallelism relies on storing the full batch sampling result in host memory, which exceeds the 512 GB system memory. SIMPLE's dynamic cache updates also introduce substantial overhead on large graphs; on the BITCOIN dataset for TGAT, its performance is even worse than TGL's. In contrast, \textsc{FAST} maintains robust execution comparable to TGL while achieving substantial speedups, reflecting the scalability and stability of our design. We further analyze the overhead characteristics of these frameworks in Section~\ref{sec:exp_overhead}.}

\subsection{\xin{Ablation Study}}
\xin{We conduct a detailed ablation study to quantify the contribution of each of FAST's three core optimizations: SlimCache(SC), thread-efficient(TE) operators, and topology-aware(TA) sampling. This analysis isolates their individual impact on performance.}

\xin{\textbf{Effectiveness of  SlimCache.} To quantify the contribution of our memory-I/O optimizations, we compare SlimCache with TASER and SIMPLE under varying cache ratios, where the cache ratio is the fraction of node and edge features stored in GPU memory (e.g., 0.2 indicates 20\% of features cached). All three frameworks share the same memory overhead $M$ as the cache budget: TASER uses a node-prioritized caching strategy, SIMPLE employs a dynamic node‑prioritized placement, and FAST applies the greedy selection-based SlimCache strategy that combines caching and compression.}

\begin{figure}
  \includegraphics[width=\columnwidth]{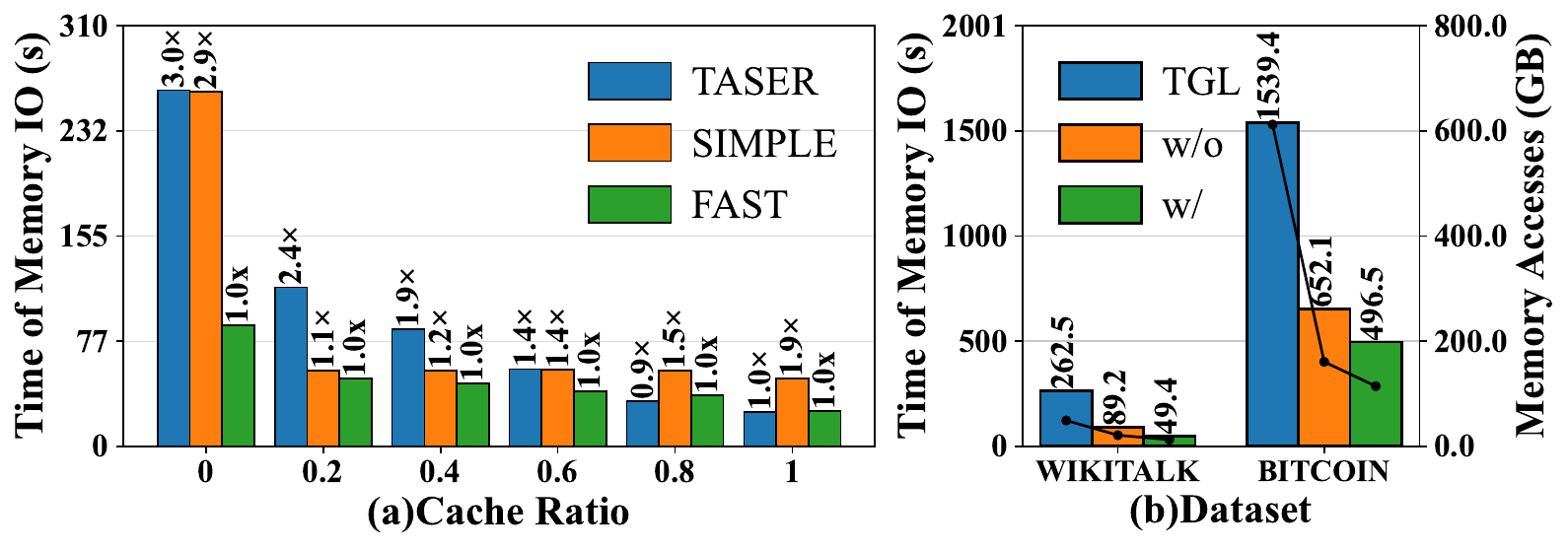}
  \caption{The time spent on the memory IO comparison between (a) TASER, SIMPLE and FAST of TGAT on WIKITALK; (b) with and without the greedy selection strategy on TGAT.}
  \label{fig:cache_access}
\end{figure}

\yushu{Figure~\ref{fig:cache_access}(a) shows memory I/O time for the TGAT model on the WIKITALK dataset. With no cache, SlimCache’s compression speeds up 3.0$\times$. At cache ratio 0.8, FAST underperforms TASER as compression overhead dominates when the cache saturates. At cache ratio 1, SlimCache automatically disables compression for small graphs (e.g., 5.74 GB features) on typical server GPUs. For large graphs such as BITCOIN (94 GB), the cache ratio must remain small, and SlimCache delivers the best memory I/O acceleration.}

\xin{Further isolating the effects of compression and caching, we set the cache ratio to 0.2 on WIKITALK and BITCOIN (Figure~\ref{fig:cache_access}(b)). Here, \texttt{w/o} denotes disabling the greedy cache (only compression), while \texttt{w} corresponds to full SlimCache. On WIKITALK, compression alone reduces I/O from 49 GB to 21 GB; adding greedy caching cuts traffic further to 13 GB, yielding a 5.3$\times$ I/O speedup. On BITCOIN, compression reduces 612 GB to 160 GB, and SlimCache brings it down to 114 GB, achieving a 3.1$\times$ I/O speedup. This confirms that compression alone improves I/O, and SlimCache’s combined design further reduces traffic under typical memory constraints.}

\begin{figure}[t]
    \centering
    \begin{minipage}{0.48\columnwidth}
        \centering
        \includegraphics[width=\linewidth]{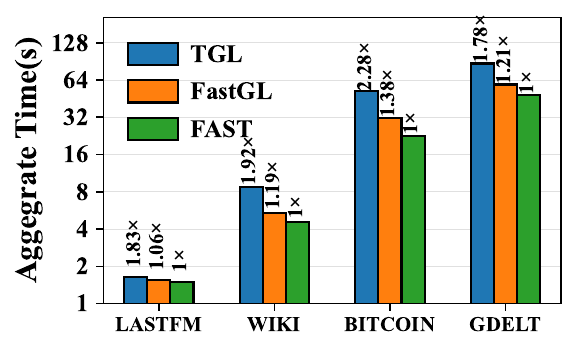}
        \\
        {\small (a)}
        \label{fig:exp_agg_spd}
    \end{minipage}
    \hfill
    \begin{minipage}{0.48\columnwidth}
        \centering
        \includegraphics[width=\linewidth]{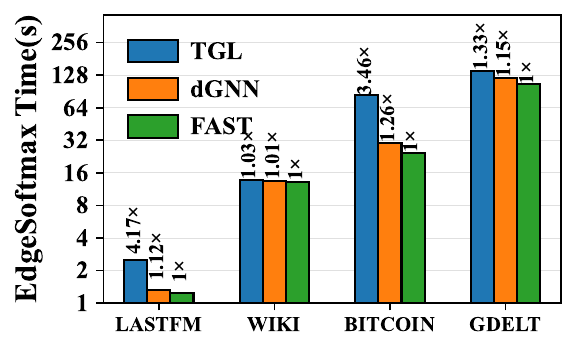}
        \\
        {\small (b)}
        \label{fig:exp_esm_spd}
    \end{minipage}
    \caption{The time spend on graph operator in TGAT. (a) TGL, FastGL and FAST of AGG;(b) TGL, dGNN and FAST of ESM.}
    \label{fig:gop_spdup}
\end{figure}

\xin{\textbf{Effectiveness of Thread-Efficient Graph Operators.} Figure~\ref{fig:gop_spdup}(a) and (b) report the graph operator execution time for DGL, dGNN, FastGL, and FAST across four datasets. FAST achieves up to 2.3$\times$ speedup for the aggregate operator and 4.2$\times$ for edge softmax.}

\xin{For aggregation, FAST achieves 2.28$\times$ speedup over TGL and 1.38$\times$ over FastGL on BITCOIN, whose high imbalance (Table~\ref{tab:moti_degree_unblance}) matches our edge-parallel COO design. For edge softmax, FAST attains 4.17$\times$ and 3.46$\times$ speedup over TGL on LASTFM and BITCOIN, respectively, benefiting from the small-degree pattern and thread-level reduction outlined in Section~\ref{sec:design_compute}. On WIKITALK, acceleration is modest due to its exponential degree distribution, which limits the benefit of the CSR-based reduction and prevents meaningful gains from COO-based trials.}

\xin{Table~\ref{tab:kernel_performance} profiles the kernel metrics for the largest sampled subgraph. For ESM, the L1 hit rate rises from 32.46\% to 90.26\%, and the L2 hit rate from 68.63\% to 80.67\%. FAST increases the nodes per block from 32  to 512, exposing additional locality and reducing access latency for the memory-bound ESM operator. The average active warps per SM rise from 12.95 to 29.58, while the average active threads per warp drop from 26 to 19, reflecting reduced invalid work rather than lower efficiency. For AGG, hit rates and active warps per SM decline slightly, as the edge-parallel scheme weakens spatial locality, but the average active threads per warp increase, directly alleviating the load imbalance in node-centric aggregation.}

\begin{table}[htbp]
  \centering
  \caption{Kernel performance comparison}
  \label{tab:kernel_performance}
  \begin{tabular}{lcccc}
    \toprule
    Kernel & L1 (\%) & L2 (\%) & Act. warps & Avg. threads \\
    \midrule
    ESM-dGNN  & 32.46 & 68.63 & 12.95 / SM & 25.84 / Warp \\
    ESM-FAST  & 90.26 & 80.67 & 29.58 / SM & 19.00 / Warp \\
    AGG-FastGL & 81.04 & 51.44 & 39.83 / SM & 20.36 / Warp \\
    AGG-FAST  & 25.99 & 46.75 & 28.66 / SM & 27.00 / Warp \\
    \bottomrule
  \end{tabular}
\end{table}

\begin{figure}
  \includegraphics[width=\columnwidth]{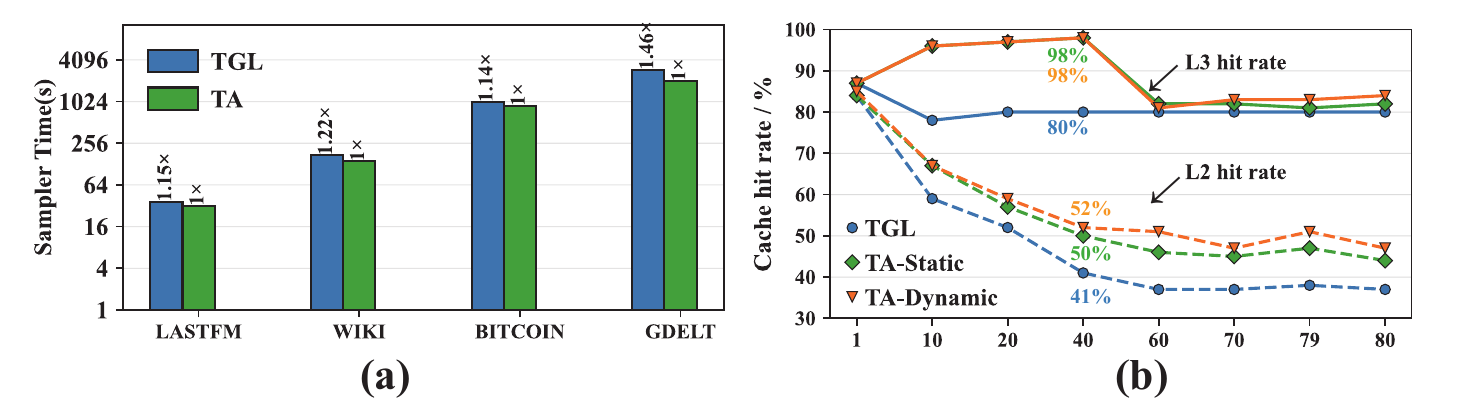}
  \caption{(a) The sampling time of TGL and FAST.(b) The cache hit rates(TGL and TA with static or dynamic binding).}
  \label{fig:time_ta}
\end{figure}

\begin{figure}
  \includegraphics[width=\columnwidth]{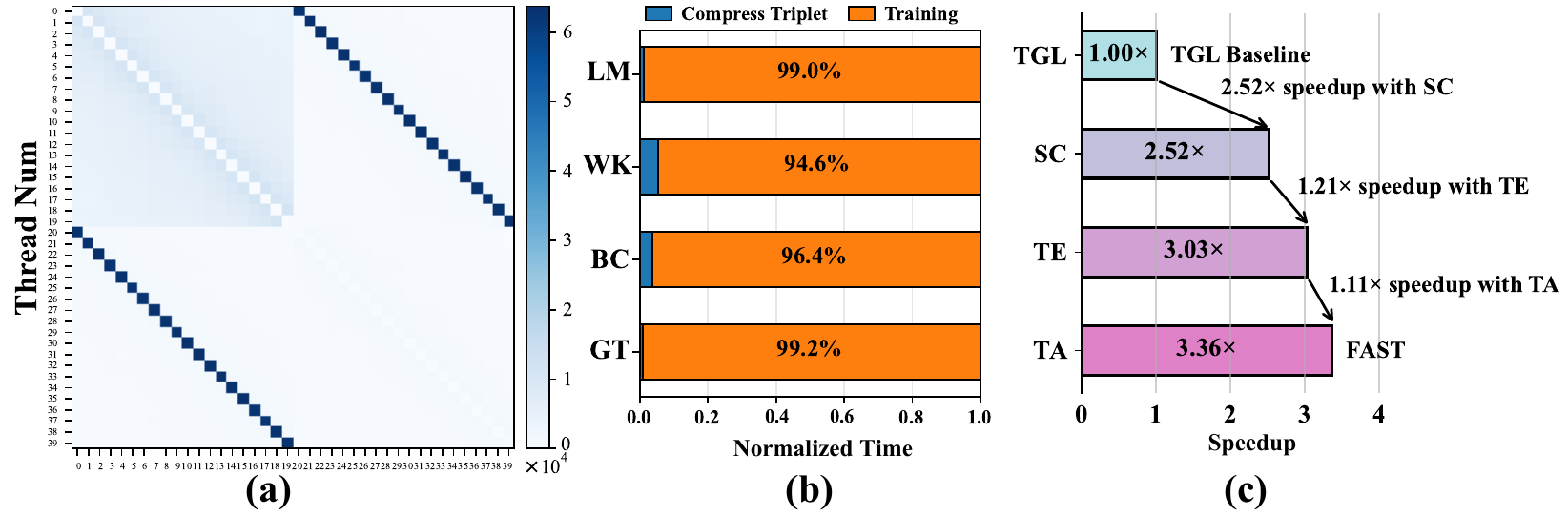}
  \caption{(a) Thread affinity matrix under 8 threads on WIKITALK.(b) Comparison of normalized time between the compressing and the training.(c) The breakdown analysis on the average overall speedup. ‘SC' denotes SlimCache, ‘TE’ denotes thread-efficient, and ‘TA’ denotes topology-aware.}
  \label{fig:exp_ct_abla}
\end{figure}

\yushu{\textbf{Topology-Aware Sampling Effectiveness.} FAST speeds sampling by 1.14$\times$–1.46$\times$ versus TGL (Figure~\ref{fig:time_ta}(a)). On WIKITALK with 40 threads, TA improves L2 cache hit rates from 41\% to 50–52\% and L3 hit rates from 80\% to 98\%(Figure~\ref{fig:time_ta}(b)). Gains stem from mapping highly affine threads to shared L2/L3 domains, as visualized in the affinity matrices of Figure~\ref{fig:exp_ct_abla}(a) (\yushu{perform thread binding based on the darkest block}). When exceeding 40 threads, L3 hit rates decline due to NUMA(Non-Uniform Memory Access) spill, yet TA outperforms TGL by binding high-affinity threads preferentially within the same NUMA node. After sampling, the compression engine outputs compressed triples that drive SlimCache and CSR-Based operator(Section~\ref{sec:impl_ct}).Figure~\ref{fig:exp_ct_abla}(b) shows that the engine’s overhead is modest, consuming 0.8\%–5.4\% of end-to-end training time (average 2.7\%) across datasets.}

\yushu{\textbf{End-to-End Speedup Breakdown.} Figure~\ref{fig:exp_ct_abla}(c) decomposes the overall speedup of TGN on four datasets versus TGL. SlimCache cuts memory I/O time and yields the largest gain. Adding TE graph operators provides an additional 1.21$\times$ speedup by tackling reduction inefficiency in ESM and load imbalance in AGG. The TA sampler then contributes about 1.1$\times$ speedup by improving cache locality in sampling. Together, the three components explain the end‑to‑end acceleration reported in Section~\ref{sec:exp_overall}.}

\subsection{Overhead}
\label{sec:exp_overhead}

\yushu{\textsc{FAST} delivers acceleration with small, manageable memory and preprocessing overhead. Evaluated on TGN with the million-scale BITCOIN dataset(Table~\ref{tab:exp_overhead}), \textsc{FAST} matches the TGL baseline in main memory usage, while SIMPLE incur OOM due to aggressive optimizations. SWIFT minimizes main memory by offloading to disk, but consumes 90$\times$ more disk space than TGL. In contrast, \textsc{FAST}'s disk usage nearly equals TGL's, as SlimCache only stores a compact hot-ID list without retaining intermediate data. GPU memory utilization is comparable to SWIFT, confirming that our caching strategy fully exploits GPU memory without storage bottlenecks.}

\yushu{Table~\ref{tab:exp_overhead} also reports the pre-sampling time.\textsc{FAST}'s overhead is 2.3$\times$ and 4.9$\times$ lower than SIMPLE and SWIFT, while TGL/ETC have none. SIMPLE uses full-batch sampling for disk-based caching; SWIFT builds disk buckets for async-I/O. \textsc{FAST} instead performs lightweight pre-sampling: topology-aware analysis generates thread-binding, and sampled subgraphs update hot lists before being discarded, keeping preprocessing minimal.}

\yushu{We further evaluate SWIFT’s per‑epoch performance, its performance is limited by 1.7 GB/s disk bandwidth on our hardware, below the reported 2.0–6.7 GB/s. This constraint prevents SWIFT from achieving the expected throughput, making \textsc{FAST} becomes 1.3$\times$ faster than SWIFT. \textsc{FAST} builds on the main-memory framework\citep{tgl,etc,simple} without new hardware constraints, demonstrating practical robustness.}

\begin{table}[t]
\centering
\small
\caption{Overhead comparison of Memory and Time.}
\label{tab:exp_overhead}
\begin{threeparttable}
\begin{tabular}{@{}lccccc@{}}
\toprule
\multirow{2}{*}{\textbf{Method}} & \multicolumn{3}{c}{\textbf{Memory usage (GB)}} & \multicolumn{2}{c}{\textbf{Time (s)}} \\
\cmidrule(lr){2-4} \cmidrule(l){5-6}
& \textbf{Main} & \textbf{Disk} & \textbf{GPU} & \textbf{Pre-sample} & \textbf{Epoch} \\
\midrule
TGL & 101.62 & 2.84 & 10.54(26.4\%) & 0 & 6824(2.5x) \\
ETC & 274.38 & 2.84 & 10.62(26.6\%) & 0 & 4029(1.4x) \\
SIMPLE & OOM & 5.53 & N/A & 285.03(2.3x) & N/A \\
SWIFT & \textbf{28.26} & 255.62 & 38.42(96.1\%) & 615.73(4.9x) & 3747(1.3x) \\
FAST & 101.59 & 2.92 & \textbf{38.89(97.2\%)} & \textbf{125.69} & \textbf{2784} \\
\bottomrule
\end{tabular}
\end{threeparttable}
\end{table}

%% file: text/relatedWork.tex
\section{Related Work}

\textbf{Graph Feature Caching.} 

\yushu{In static GNNs, FastGL \citep{Zhu2024FastGLAG} pre-samples to identify hot features \citep{Yang2022GNNLabAF} and uses batch reordering to retain previously accessed data, combining static and dynamic caching. In TGNNs training, TASER \citep{taser} caches statically per epoch based solely on edge feature access frequencies, while SIMPLE \citep{simple} selects retention intervals for dynamic placement. ETC \citep{etc} reduces I/O volume through feature compression and overlaps GPU computation with CPU compression, masking the compression overhead. However, SIMPLE’s and ETC’s dependence on full‑batch sampled subgraph data introduces host memory bottlenecks on large graphs. SWIFT \citep{swift_2025} alleviates memory pressure via a disk‑based I/O pipeline, but its bucket‑based design needs much more disk space and is constrained by disk bandwidth.}

\textbf{Computational Optimization.}
\xin{Several TGNN works accelerate computation via reuse or redundancy elimination. Orca \cite{Li2023OrcaST} caches embeddings, while TGOpt \cite{Wang2023TGOptRO} removes redundant computations. In static GNNs, GNNOne \citep{GNNone} optimizes DGL’s core kernels (SpMM and SDDMM) via a two‑phase data loading and reuse strategy; dGNN \citep{dGNN} and FuseGNN \citep{fuseGNN} reorder and fuse operator for full‑graph training, and FastGL \citep{Zhu2024FastGLAG} caches edge features to accelerate aggregation. These designs are tailored to static GNNs and are not directly applicable to highly sparse TGNN workloads.}

\textbf{Sampler-Based Acceleration.}
\xin{TGL \citep{tgl} is a general framework for training TGNNs on large continuous‑time dynamic graphs, proposing T-CSR and CPU temporal sampler.Subsequent works exploit GPU acceleration for specific scenarios, include TASER \cite{taser} (adaptive), GNNFlow \cite{gnnflow} (distributed), MSpipe \cite{msPipe} (memory‑based), SWIFT \cite{swift_2025} (UVA). These GPU samplers are often highly customized and lack easy reuse and clean cross‑framework comparisons.}

%% file: text/conclusion.tex
\section{Conclusion}

\xin{In this paper, we propose \textsc{FAST}, a holistic framework for optimizing memory I/O, computation, and sampling in Temporal GNN training on large dynamic graphs. By jointly exploiting within‑batch and cross‑batch redundancy, \textsc{FAST} reduces host–device data movement, accelerates sparse graph operators, and improves CPU cache locality in temporal neighbor sampling. Extensive experiments show that \textsc{FAST} achieves an average 2.1$\times$ speedup (up to 4.7$\times$) over state‑of‑the‑art systems without sacrificing model accuracy, demonstrating that a cross‑stage co‑design can significantly improve TGNN training.}

%% file: references.bib
@String{Computing = "Computing" }

@String{Computer = "{IEEE} Computer" }

@article{simple,
author = {Gao, Shihong and Li, Yiming and Zhang, Xin and Shen, Yanyan and Shao, Yingxia and Chen, Lei},
title = {SIMPLE: Efficient Temporal Graph Neural Network Training at Scale with Dynamic Data Placement},
year = {2024},
issue_date = {June 2024},
publisher = {Association for Computing Machinery},
address = {New York, NY, USA},
volume = {2},
number = {3},
url = {https://doi.org/10.1145/3654977},
doi = {10.1145/3654977},
journal = {Proc. ACM Manag. Data},
month = may,
articleno = {174},
numpages = {25}
}

@inproceedings{nextDoor_Jangda2020AcceleratingGS,
author = {Jangda, Abhinav and Polisetty, Sandeep and Guha, Arjun and Serafini, Marco},
title = {Accelerating graph sampling for graph machine learning using GPUs},
year = {2021},
isbn = {9781450383349},
publisher = {Association for Computing Machinery},
address = {New York, NY, USA},
url = {https://doi.org/10.1145/3447786.3456244},
doi = {10.1145/3447786.3456244},
booktitle = {Proceedings of the Sixteenth European Conference on Computer Systems},
pages = {311–326},
numpages = {16},
location = {Online Event, United Kingdom},
series = {EuroSys '21}
}

@article{etc,
author = {Gao, Shihong and Li, Yiming and Shen, Yanyan and Shao, Yingxia and Chen, Lei},
title = {ETC: Efficient Training of Temporal Graph Neural Networks over Large-Scale Dynamic Graphs},
year = {2024},
issue_date = {January 2024},
publisher = {VLDB Endowment},
volume = {17},
number = {5},
issn = {2150-8097},
url = {https://doi.org/10.14778/3641204.3641215},
doi = {10.14778/3641204.3641215},
journal = {Proc. VLDB Endow.},
month = jan,
pages = {1060–1072},
numpages = {13}
}

@inproceedings{sankar2020dysat,
author = {Sankar, Aravind and Wu, Yanhong and Gou, Liang and Zhang, Wei and Yang, Hao},
title = {DySAT: Deep Neural Representation Learning on Dynamic Graphs via Self-Attention Networks},
year = {2020},
isbn = {9781450368223},
publisher = {Association for Computing Machinery},
address = {New York, NY, USA},
url = {https://doi.org/10.1145/3336191.3371845},
doi = {10.1145/3336191.3371845},
booktitle = {Proceedings of the 13th International Conference on Web Search and Data Mining},
pages = {519–527},
numpages = {9},
location = {Houston, TX, USA},
series = {WSDM '20}
}

@inproceedings{GNNOne,
author = {Gong, Yidong and Kumar, Pradeep},
title = {GNNOne: A Unified System Optimizations for GNN Kernels},
year = {2024},
isbn = {9798400704130},
publisher = {Association for Computing Machinery},
address = {New York, NY, USA},
url = {https://doi.org/10.1145/3625549.3658655},
doi = {10.1145/3625549.3658655},
booktitle = {Proceedings of the 33rd International Symposium on High-Performance Parallel and Distributed Computing},
pages = {15–27},
numpages = {13},
location = {Pisa, Italy},
series = {HPDC '24}
}

@inproceedings{ht_icpp,
  author       = {Konstantin Macarenco and
                  Kristina Frye and
                  Benjamin Hamlin and
                  Karen L. Karavanic},
  title        = {The Effects of System Management Interrupts on Multithreaded, Hyper-threaded,
                  and {MPI} Applications},
  booktitle    = {45th International Conference on Parallel Processing Workshops, {ICPP}
                  Workshops 2016, Philadelphia, PA, USA, August 16-19, 2016},
  pages        = {338--345},
  publisher    = {{IEEE} Computer Society},
  year         = {2016},
  url          = {https://doi.org/10.1109/ICPPW.2016.55},
  doi          = {10.1109/ICPPW.2016.55},
  bibsource    = {dblp computer science bibliography, https://dblp.org}
}

@article{swift_2025,
author = {Guo, Rui and Ding, Zezhong and Xie, Xike and Xu, Jianliang},
title = {SWIFT: Enabling Large-Scale Temporal Graph Learning on a Single Machine},
year = {2025},
issue_date = {September 2025},
publisher = {Association for Computing Machinery},
address = {New York, NY, USA},
volume = {3},
number = {4},
url = {https://doi.org/10.1145/3749184},
doi = {10.1145/3749184},
journal = {Proc. ACM Manag. Data},
month = sep,
articleno = {266},
numpages = {27}
}

@inproceedings{Zhu2024FastGLAG,
author = {Zhu, Zeyu and Wang, Peisong and Hu, Qinghao and Li, Gang and Liang, Xiaoyao and Cheng, Jian},
title = {FastGL: A GPU-Efficient Framework for Accelerating Sampling-Based GNN Training at Large Scale},
year = {2025},
isbn = {9798400703911},
publisher = {Association for Computing Machinery},
address = {New York, NY, USA},
url = {https://doi.org/10.1145/3622781.3674167},
doi = {10.1145/3622781.3674167},
booktitle = {Proceedings of the 29th ACM International Conference on Architectural Support for Programming Languages and Operating Systems, Volume 4},
pages = {94–110},
numpages = {17},
location = {Hilton La Jolla Torrey Pines, La Jolla, CA, USA},
series = {ASPLOS '24}
}

@inproceedings{Lo2021EGraphSAGEAG,
author = {Liu, Yuwen and Qi, Lianyong and Liu, Weiming and Xu, Xiaolong and Zhang, Xuyun and Dou, Wanchun},
title = {GraphSAGE-based POI Recommendation via Continuous-Time Modeling},
year = {2024},
isbn = {9798400701726},
publisher = {Association for Computing Machinery},
address = {New York, NY, USA},
url = {https://doi.org/10.1145/3589335.3651515},
doi = {10.1145/3589335.3651515},
booktitle = {Companion Proceedings of the ACM Web Conference 2024},
pages = {585–588},
numpages = {4},
location = {Singapore, Singapore},
series = {WWW '24}
}

@article{Chen2023NeutronStreamAD,
author = {Chen, Chaoyi and Gao, Dechao and Zhang, Yanfeng and Wang, Qiange and Fu, Zhenbo and Zhang, Xuecang and Zhu, Junhua and Gu, Yu and Yu, Ge},
title = {NeutronStream: A Dynamic GNN Training Framework with Sliding Window for Graph Streams},
year = {2023},
issue_date = {November 2023},
publisher = {VLDB Endowment},
volume = {17},
number = {3},
issn = {2150-8097},
url = {https://doi.org/10.14778/3632093.3632108},
doi = {10.14778/3632093.3632108},
journal = {Proc. VLDB Endow.},
month = nov,
pages = {455–468},
numpages = {14}
}

@inproceedings{icpp_openmp,
author = {Yu, Chenle and Royuela, Sara and Qui\~{n}ones, Eduardo},
title = {Enhancing Heterogeneous Computing Through OpenMP and GPU Graph},
year = {2024},
isbn = {9798400717932},
publisher = {Association for Computing Machinery},
address = {New York, NY, USA},
url = {https://doi.org/10.1145/3673038.3673050},
doi = {10.1145/3673038.3673050},
booktitle = {Proceedings of the 53rd International Conference on Parallel Processing},
pages = {534–543},
numpages = {10},
location = {Gotland, Sweden},
series = {ICPP '24}
}

@inproceedings{msPipe,
author = {Sheng, Guangming and Su, Junwei and Huang, Chao and Wu, Chuan},
title = {MSPipe: Efficient Temporal GNN Training via Staleness-Aware Pipeline},
year = {2024},
isbn = {9798400704901},
publisher = {Association for Computing Machinery},
address = {New York, NY, USA},
url = {https://doi.org/10.1145/3637528.3671844},
doi = {10.1145/3637528.3671844},
booktitle = {Proceedings of the 30th ACM SIGKDD Conference on Knowledge Discovery and Data Mining},
pages = {2651–2662},
numpages = {12},
location = {Barcelona, Spain},
series = {KDD '24}
}

@inproceedings{fuseGNN,
author = {Chen, Zhaodong and Yan, Mingyu and Zhu, Maohua and Deng, Lei and Li, Guoqi and Li, Shuangchen and Xie, Yuan},
title = {fuseGNN: accelerating graph convolutional neural network training on GPGPU},
year = {2020},
isbn = {9781450380263},
publisher = {Association for Computing Machinery},
address = {New York, NY, USA},
url = {https://doi.org/10.1145/3400302.3415610},
doi = {10.1145/3400302.3415610},
booktitle = {Proceedings of the 39th International Conference on Computer-Aided Design},
articleno = {60},
numpages = {9},
location = {Virtual Event, USA},
series = {ICCAD '20}
}

@inproceedings{Yang2022GNNLabAF,
author = {Yang, Jianbang and Tang, Dahai and Song, Xiaoniu and Wang, Lei and Yin, Qiang and Chen, Rong and Yu, Wenyuan and Zhou, Jingren},
title = {GNNLab: a factored system for sample-based GNN training over GPUs},
year = {2022},
isbn = {9781450391627},
publisher = {Association for Computing Machinery},
address = {New York, NY, USA},
url = {https://doi.org/10.1145/3492321.3519557},
doi = {10.1145/3492321.3519557},
booktitle = {Proceedings of the Seventeenth European Conference on Computer Systems},
pages = {417–434},
numpages = {18},
location = {Rennes, France},
series = {EuroSys '22}
}

@article{Li2023OrcaST,
author = {Li, Yiming and Shen, Yanyan and Chen, Lei and Yuan, Mingxuan},
title = {Orca: Scalable Temporal Graph Neural Network Training with Theoretical Guarantees},
year = {2023},
issue_date = {May 2023},
publisher = {Association for Computing Machinery},
address = {New York, NY, USA},
volume = {1},
number = {1},
url = {https://doi.org/10.1145/3588737},
doi = {10.1145/3588737},
journal = {Proc. ACM Manag. Data},
month = may,
articleno = {52},
numpages = {27}
}

@inproceedings{Wang2023TGOptRO,
author = {Wang, Yufeng and Mendis, Charith},
title = {TGOpt: Redundancy-Aware Optimizations for Temporal Graph Attention Networks},
year = {2023},
isbn = {9798400700156},
publisher = {Association for Computing Machinery},
address = {New York, NY, USA},
url = {https://doi.org/10.1145/3572848.3577490},
doi = {10.1145/3572848.3577490},
booktitle = {Proceedings of the 28th ACM SIGPLAN Annual Symposium on Principles and Practice of Parallel Programming},
pages = {354–368},
numpages = {15},
location = {Montreal, QC, Canada},
series = {PPoPP '23}
}

@inproceedings{dgl,
  title = {Deep graph library: Towards efficient and scalable deep learning on graphs},
  author = {Minjie Yu Wang},
  booktitle = {ICLR workshop on representation learning on graphs and manifolds},
  year = {2019},
  note = {doi:\url{https://doi.org/10.48550/arXiv.1909.01315}}
}

@inproceedings{tgn,
  author = {Rossi, Emanuele and Chamberlain, Ben and Frasca, Fabrizio and Eynard, Davide and Monti, Federico and Bronstein, Michael},
  title = {Temporal Graph Networks for Deep Learning on Dynamic Graphs},
  booktitle = {Proceedings of the ICML 2020 Workshop on Graph Representation Learning},
  year = {2020}
}

@inproceedings{tgat,
  title={Inductive representation learning on temporal graphs},
  author={Xu, Da and Ruan, Chuanwei and K{\"o}rpeo{\u{g}}lu, Evren and Kumar, Sushant and Achan, Kannan},
  booktitle={International Conference on Learning Representations (ICLR)},
  year={2020}
}

@article{tgl,
  title={TGL: A General Framework for Temporal GNN Training onBillion-Scale Graphs},
  author={Hongkuan Zhou and Da Zheng and Israt Nisa and Vasileios Ioannidis and Xiang Song and George Karypis},
  journal={Proc. VLDB Endow.},
  year={2022},
  volume={15}, 
  number={8},
  pages={1572-1580}
}

@inproceedings{Jin2023SpatioTemporalGN,
  title={Spatio-Temporal Graph Neural Point Process for Traffic Congestion Event Prediction},
  author={G. Jin and Lingbo Liu and Fuxian Li and Jincai Huang},
  journal={AAAI},
  year={2023},
  volume={abs/2311.08635},
  pages={14268-14276},
}

@article{Zhang2021DynamicGN,
author = {Zhang, Mengqi and Wu, Shu and Yu, Xueli and Liu, Qiang and Wang, Liang},
title = {Dynamic Graph Neural Networks for Sequential Recommendation},
year = {2023},
issue_date = {May 2023},
publisher = {IEEE Educational Activities Department},
address = {USA},
volume = {35},
number = {5},
issn = {1041-4347},
url = {https://doi.org/10.1109/TKDE.2022.3151618},
doi = {10.1109/TKDE.2022.3151618},
journal = {IEEE Trans. on Knowl. and Data Eng.},
month = may,
pages = {4741–4753},
numpages = {13}
}

@article{bitcoin1_kondor2014rich,
	title = {Do the {Rich} {Get} {Richer}? {An} {Empirical} {Analysis} of the {Bitcoin} {Transaction} {Network}},
	volume = {9},
	url = {https://doi.org/10.1371/journal.pone.0086197},
	doi = {10.1371/journal.pone.0086197},
	number = {2},
	journal = {PLOS ONE},
	publisher = {Public Library of Science},
	author = {Kondor, Dániel and Pósfai, Márton and Csabai, István and Vattay, Gábor},
	month = feb,
	year = {2014},
	pages = {1--10},
}

@article{gnnflow,
  title={GNNFlow: A Distributed Framework for Continuous Temporal GNN Learning on Dynamic Graphs},
  author={Zhong, Yuchen and Sheng, Guangming and Qin, Tianzuo and Wang, Minjie and Gan, Quan and Wu, Chuan},
  year={2023},
  eprint={2311.17410},
  archivePrefix={arXiv},
  primaryClass={cs.DC}
}

@inproceedings{dGNN,
  title = {Understanding GNN Computational Graph: A Coordinated Computation, IO, and Memory Perspective},
  author = {Zhang, Hengrui and Yu, Zhongming and Dai, Guohao and Huang, Guyue and Ding, Yufei and Xie, Yuan and Wang, Yu},
  booktitle = {Proceedings of Machine Learning and Systems (MLSys)},
  year = {2022},
  volume = {4},
  pages = {467--484},
  url = {https://proceedings.mlsys.org/paper/2022/file/b559156047e50cf316207249d0b5a6c5-Paper.pdf},
  doi = {10.48550/arXiv.2110.09524}
}

@inproceedings{taser,
  author={Deng, Gangda and Zhou, Hongkuan and Zeng, Hanqing and Xia, Yinglong and Leung, Christopher and Li, Jianbo and Kannan, Rajgopal and Prasanna, Viktor},
  booktitle={2024 IEEE International Parallel and Distributed Processing Symposium (IPDPS)}, 
  title={TASER: Temporal Adaptive Sampling for Fast and Accurate Dynamic Graph Representation Learning}, 
  year={2024},
  volume={},
  number={},
  pages={926-937},
  doi={10.1109/IPDPS57955.2024.00087}}

@article{milakov2018online_safe,
  title = {Online Normalizer Calculation for Softmax},
  author = {Milakov, Maxim and Gimelshein, Natalia},
  year = {2018},
  eprint = {1805.02867},
  archivePrefix = {arXiv}
}

@inproceedings{gdelt_leetaru2013gdelt,
  title={Gdelt: Global data on events, location, and tone, 1979--2012},
  author={Leetaru, Kalev and Schrodt, Philip A},
  booktitle={ISA annual convention},
  volume={2},
  number={4},
  pages={1--49},
  year={2013},
  organization={Citeseer}
}

@inproceedings{bitcoin2_rossi2015network,
  title={The network data repository with interactive graph analytics and visualization},
  author={Rossi, Ryan and Ahmed, Nesreen},
  booktitle={Proceedings of the AAAI conference on artificial intelligence},
  volume={29},
  number={1},
  year={2015}
}

@inproceedings{wiki2_paranjape2017motifs,
  title={Motifs in temporal networks},
  author={Paranjape, Ashwin and Benson, Austin R and Leskovec, Jure},
  booktitle={Proceedings of the tenth ACM international conference on web search and data mining},
  pages={601--610},
  year={2017}
}

@inproceedings{lastfm_kumar2019predicting,
  title={Predicting dynamic embedding trajectory in temporal interaction networks},
  author={Kumar, Srijan and Zhang, Xikun and Leskovec, Jure},
  booktitle={Proceedings of the 25th ACM SIGKDD international conference on knowledge discovery \& data mining},
  pages={1269--1278},
  year={2019}
}

@article{blossom_algo_shoemaker2016edmonds,
  title={Edmonds’ blossom algorithm},
  author={Shoemaker, Amy and Vare, Sagar},
  journal={CME},
  volume={18},
  year={2016}
}
